\documentclass[journal]{IEEEtran}
\usepackage[numbers,sort&compress]{natbib}
\usepackage{graphicx}
\usepackage{algorithmic}
\usepackage{algorithm}
\usepackage{multirow}
\usepackage{multicol}
\ifCLASSINFOpdf
\else
\fi
\usepackage{amsmath}
\begin{document}
%
\title{Importance Filtered Cross-Domain Adaptation}
%
%
%
\author{Wei~Wang,
		Haojie~Li$^\ast$,
        Zhihui~Wang,
        Jing~Sun,
        Zhengming~Ding,
        and~Fuming~Sun
\thanks{W. Wang, H. Li, Z. Wang, and J. Sun were with the DUT-RU International School of Information Science \& Engineering, Dalian University of Technology, Dalian, Liaoning, 116000, P.R. China e-mail: (WWLoveTransfer@mail.dlut.edu.cn, hjli@dlut.edu.cn, zhwang@dlut.edu.cn, sunjing616@mail.dlut.edu.cn).}
\thanks{Z. Ding was with the Department of Computer, Information and Technology, Purdue School of Engineering and Technology, Indiana University-Purdue University Indianapolis, Indianapolis, IN, 46202, USA e-mail: zd2@iu.edu.cn}
\thanks{F. Sun was with the School of Information \& Communication Engineering, Dalian Minzu University of Technology, Dalian, Liaoning, 116000, P.R. China e-mail: sunfuming@dlnu.edu.cn.}}	

%
%

\markboth{IEEE TRANSACTIONS ON **}%
{Wang \MakeLowercase{\textit{et al.}}: Bare Demo of IEEEtran.cls for IEEE Journals}
%



\maketitle

\begin{abstract}
In Domain Adaptation (DA), the category-relevant losses usually occupy a dominant position, while they are usually built with hard or soft labels in existing models. We observed that hard labels are overconfident due to hard samples existed, and soft labels are ambiguous as too many small noisy probabilities involved, and both of them are easily to cause negative transfer. Besides, the category-irrelevant losses in Closed-Set DA (CSDA) paradigm fail to work in Open-Set DA (OSDA), and they also have to be in a category-relevant form, since target data samples are split into shared and private classes. To this end, we propose a newly-unified DA framework (i.e., Importance Filtered Cross-Domain Adaptation, IFCDA). Firstly, an importance filtered mechanism is devised to generate filtered soft labels to mitigate negative transfer desirably. Specifically, the soft labels are divided into confident and ambiguous ones. Then, only the maximum probability in each confident label is retained, and a threshold value is set to truncate each ambiguous label so that only prominent probabilities are reserved. Moreover, a general graph-based label propagation is contrived to attain soft labels in both CSDA and OSDA, where an extra component is embedded into label vector, so that it could detect target novel classes. Finally, the category-relevant losses in both scenarios are reformulated using filtered soft labels, while the category-irrelevant MMD loss in CSDA is reformulated as a form like class-wise MMD using newly-designed importance filtered soft labels. Notably, CSDA paradigm is a special case when all extra components are set to 0, thus the proposed approach is geared to both CSDA and OSDA. Comprehensive experiments on benchmark cross-domain object recognition datasets verify that the proposed approach outperforms several state-of-the-art methods in both scenarios.

\end{abstract}


%
\IEEEpeerreviewmaketitle

\section{Introduction}
%
%
%
%
\IEEEPARstart{A}n assumption of traditional machine learning algorithms is that the training (i.e., source domain) and test (i.e., target domain) data are drawn from the same or similar distribution, while it will fail to hold in complicated practical applications. As we all know, it is impractical to relabel a large amount of data for each coming domain since it is time-consuming and labor-intensive. Fortunately, Domain Adaptation (DA) has been proposed to address this challenge, and its goal is to employ previously labeled source domain to boost the task in a newly unlabeled target domain. It has been widely applied to cross-domain image classification \cite{IEEE:DICD}, depth estimation\cite{IEEE:DA_DE1} and semantic segmentation \cite{IEEE:DA_SS1}, etc. 

\par The most extensively studied DA technique is the Feature-based Domain Adaptation (FDA), which aims to leverage cross domain-invariant features and specific domain-invariant structures with feature transformation, while geometrical and statistical information of data is further exploited. Gopalan et al. \cite{IEEE:SGF} proposed an approach of Sampling Geodesic Flow (SGF) which samples a fixed number of subspaces from a geometrical flow curve, and projects original features into these subspaces, then concatenates them as new feature representations. Gong et al. \cite{IEEE:GFK} further proposed Geodesic Flow Kernel (GFK) to integrate an infinite number of subspaces, to derive their feature representations with infinite dimensions. Fernando et al. \cite{IEEE:SA} presented a method of Subspace Alignment (SA), where the source and target subspaces can be aligned by computing the linear projection that minimizes the Frobenius norm of the difference between those two subspaces. 

\par However, relying solely on the geometrical transformation is insufficient for cross domain-invariant features. To this end, Sun et al. \cite{IEEE:SDA} presented a unified view of subspace mapping-based methods, and a generalize approach that aligns the distributions as well as the subspaces (i.e., Subspace Distribution Alignment, SDA). In addition, since domain-specific structures of data would be degraded during transferable feature learning \cite{IEEE:BSP}, Long et al. \cite{IEEE:GTL} proposed a method of  Graph co-regularized Transfer Learning (GTL) to not only align the source and target subspaces, but also preserve local geometrical structure of data manifold. Ghifary et al. \cite{IEEE:SCA} raised a unified framework for DA (i.e., Scatter Component Analysis, SCA), with discriminative structure of source domain being respected.

\par From the above discussion, it is apparent that the FDA methods mainly make efforts on how to devise geometrical and statistical losses effectively and robustly, to not only prompt cross domain-invariant features, but also preserve domain-specific structures hidden in original data space (i.e., specific domain-invariant structures). It is generally known that geometrical and statistical losses could be divided into category-relevant and category-irrelevant ones. For example, the Laplacian regularization term \cite{IEEE:LEST} aimed to exploit the local geometrical structure of data manifold, and the Maximum Mean Discrepancy (MMD) \cite{IEEE:MMD} and the wasserstein distance \cite{IEEE:OPT} were modeled to measure the marginal distribution difference between the source and target domains, all of them pertain to category-irrelevant losses due to no labels needed. As for the category-relevant losses, the class scatter matrix in linear discriminant analysis was constructed to respect the discriminative structure of data \cite{IEEE:DICD}, and the class-wise MMD \cite{IEEE:JDA} was devised to decrease the conditional distribution divergence across different domains.   

\par Recent remarkable DA approaches \cite{IEEE:DICD,IEEE:LPJT,IEEE:TIT} indicated that the performance of DA is mainly determined by category-relevant losses, thus how to correctly model them is a key factor in despite of no target labels, since the DA setting discussed in this paper is unsupervised and only the source labels are available. However, a shortcoming shared by most of DA methods is that the category-relevant losses are usually formulated using pseudo hard target labels, such as the class-wise MMD and class scatter matrix. Since the probabilities of data points belonging to each class given by the hard labels are either 0 or 1. This could hurt the knowledge transfer when target samples are predicted wrongly in the beginning due to hard samples existed \cite{IEEE:GAKT}. For example, when target samples from two classes have overlap distribution (i.e., hard samples), it would easily undermine the intrinsic structure of data by assigning only hard labels to those samples.

\par Differently, the soft labels provide confidence values between 0 and 1 for those probabilities, which offers the possibility for reformulating those category-relevant losses in a probabilistic weighted manner, to alleviate negative transfer incurred by hard labels. In this regard, Ding et al. \cite{IEEE:GAKT} introduced a probabilistic class-wise MMD for cross-domain distributional alignment (i.e., Graph Adaptive Knowledge Transfer, GAKT). However, the soft labels can also lead to negative transfer since the small irrelevant probabilities existed in soft labels are very confusing. As such, the DA model would be unstable as a large amount of knowledge is wrongly transferred into target domain. This is mainly stemmed from the following reason: when a data instance just hesitates in several classes or definitely belongs to one class, it is apparent that taking those probabilities all into account will introduce so many noises. 

\par All the methods discussed above are only geared to Closed-Set DA (CSDA) scenario \cite{IEEE:OSDA}, where the same label space is shared across the source and target domains. However, since we cannot decide whether source label space is the same as target one if no target annotations are available, a more realistic setting, Open-Set DA (OSDA), is recently studied. In this paper, we mainly follow the setting from Saito et al. \cite{IEEE:OSDA_B} and Liu et al. \cite{IEEE:STA_OSDA}, where the target domain has all classes in the source domain and further contains target-specific classes. Moreover, those target novel classes are usually treated as one class due to no prior knowledge about them. However, the category-irrelevant losses in CSDA would fail to directly work in OSDA, since data samples in target domain could be categorized into shared and private subsets. Thus, they also have to be reformulated as a category-relevant form. 

\subsection{Our contribution}
\par Different from them, this paper proposes a newly-unified DA framework (i.e., Importance Filtered Cross-Domain Adaptation, IFCDA), to address the challenges discussed above. First and foremost, an importance filtered mechanism is devised to generate filtered soft labels, which aims to throw away negligible confusing probabilities but prominent ones are attained in soft labels, thus negative transfer incurred by both hard and soft labels is mitigated desirably. Specifically, the soft labels could be divided into two label subsets, i.e., the confident labels and ambiguous ones. On one hand, we directly obtain the hard form of each confident label by only retaining its maximum probability. On the other hand, a threshold value is set to truncate each ambiguous label so that only prominent probabilities are reserved. To attain soft labels in both CSDA and OSDA scenarios, we devise a General Graph-based Label Propagation (GGLP) approach, where the source labels are propagated to target by establishing a neighborhood similarity graph, and an additional component is embedded into each label vector to detect target novel classes. Remarkably, the graph in GGLP is updated with transferable features leveraged in DA procedure for more accurate soft labels. In return, the learned soft labels in GGLP are refined using proposed importance filtered mechanism, to model more effective losses in DA, which could facilitate more positive knowledge transfer.

\par Moreover, we reformulate the class-wise MMD and class scatter matrix shared by CSDA and OSDA scenarios using the proposed filtered soft labels. Since the category-irrelevant MMD loss in CSDA scenario is category-relevant in OSDA, we reformulate it as a form like class-wise MMD using newly-designed importance filtered soft labels, which provide the probabilities of data points pertaining to the shared and private classes. Specifically, the newly-designed importance filtered soft labels are computed as follows: it is assumed that the probabilities of a data instance belonging to all classes (i.e., filtered soft label) are $[a_1,a_2,a_3,a_4]$, where the first three values represent the probabilities of shared classes, and the last one denotes the probability of novel classes, thus the newly-designed importance filtered soft label is $[a_1+a_2+a_3,a_4]$. Notably, all newly-defined losses in this paper are computed using the whole data samples but only the probabilities pertaining to shared classes considered, and it actually belongs to a probabilistic weighted mechanism. As such, the shared knowledge between the two different domains in OSDA paradigm could be transferred correctly, where target data instances from shared classes are assigned with larger probabilistic weights than those from private ones. Apparently, the CSDA paradigm is a special case of the proposed model when all additional components are set to 0, thus our approach could be geared to both CSDA and OSDA settings and it is more general than existing algorithms.   


\par The main contributions of our work are three-folds:

\begin{itemize}
	\item We propose an importance filtered mechanism to throw away negligible confusing probabilities but prominent ones are retained, so that negative transfer incurred by hard and soft labels could be mitigated desirably.
	\item To attain soft labels in both CSDA and OSDA scenarios, we devise a general graph-based label propagation method, where an additional component is embedded into each label vector to detect target novel classes.
	\item We propose a newly-unified framework to address the challenges of both CSDA and OSDA paradigms, where both the category-relevant and category-irrelevant losses are reformulated as more general forms using the proposed filtered soft labels.   
\end{itemize}

\par The remainder of this paper is organized as follows. In Section \ref{Related Work}, we review the related work in domain adaptation. In Section \ref{Filtered Soft Labels for Domain Adaptation}, we propose the newly-unified DA framework to deal with challenges of both CSDA and OSDA scenarios. The experimental evaluations are discussed in Section \ref{Experiments}. Finally, we conclude this paper in Section \ref{Conclusion}.
 
\section{Related Work}
\label{Related Work}


\subsection{Feature transformation Strategies}
\par SGF and GFK utilized a fixed or infinite number of projections to map the source and target original features as new feature representations. However, it is inconvenient to impregnate their frameworks with other essential losses. SA employed one projection to cast the source domain to target domain, so that their subspaces are aligned soundly, but the data properties of target could not be respected. The easiest tactic to manipulate is to only utilize one projection, to map the source and target domains into a common space, such as TCA (i.e., Transfer Component Analysis) \cite{IEEE:TCA}, TJM (i.e., Transfer Joint Matching) \cite{IEEE:TJM}, etc. However, existing outstanding DA methods have revealed that there may not exist a shared space where the distributions of two domains are the same and the data properties are also maximumly preserved in the mean time. Therefore, Zhang et al. \cite{IEEE:JGSA} first raised two projections to transform those two domains into their respective subspaces, and minimize the Frobenius norm of the difference of those two projections. This paper also utilizes two projections as they did, so that cross domain-invariant features and specific domain-invariant structures could be easily leveraged. Another representative methods with the transformation of Deep neural network (i.e., Deep FDA) have aroused recent attention \cite{IEEE:BSP,IEEE:CDAN}, while using a large amount of labeled training data does not warrant a better performance by these models due to the inherent bias within different datasets \cite{IEEE:IBDA1}. Although the proposed method is a shallow FDA paradigm, the competitive capability comparing to those Deep FDA methods has been validated on the pre-extracted deep features.  

  
\subsection{Geometrical and Statistical Alignment for Cross Domain-invariant Features}

\par What follows is how to realize geometrical and statistical calibration for cross domain-invariant features, once the transformation skeleton is established. As for geometrical alignment, existing methods usually apply different feature transformation strategies described above. With regard to statistical calibration, TCA adopted the MMD loss to diminish deviation (i.e., Frobenius norm) of means computed by source and target domains, either in Euclidean space or Reproductive Kernel Hilbert Space. Long et al. \cite{IEEE:JDA} further devised a class-wise MMD loss to narrow the means deviation in each specific class (i.e., Joint Distribution Adaptation, JDA). Courty et al. \cite{IEEE:OPT} employed the wasserstein distance loss to measure transport expenses, and hunted for an optimal transport programming to transform source domain to target (i.e., Optimal Transport for DA, OTDA). Xu et al. \cite{IEEE:LRSR} took the merit of matrix reconstruction method and represented each target samples by a combination of source samples, then the two different domains could be bridged by the low-rank and sparse structure of representation matrix. This paper also adopts the MMD and class-wise MMD since their simplicity and solid theoretical foundations, while they are reformulated as a probabilistic weighted manner with the proposed filtered soft labels. 

\subsection{Geometrical and Statistical Preservation for Specific Domain-invariant Structures}

\par Recent advances have demonstrated that the domain-specific structures of data might be degraded during domain-invariant features learning \cite{IEEE:BSP}. As such, data structures hidden in original data space also have to be respected (i.e., specific domain-invariant structures). Geometrically, MEDA (i.e., Manifold Embedding Distribution Alignment) \cite{IEEE:MEDA}, TIT (i.e., Transfer Independently Together) \cite{IEEE:TIT} and GEF (i.e., Graph Embedding Framework) \cite{IEEE:GEF} constructed a Laplacian graph to enable the embedded representations of two data points closer if they are $p$-nearest neighbors to each other, so that the local manifold structure hidden in original space could be retained. Statistically, DICD (i.e., Domain Invariant and Class Discriminative feature learning) \cite{IEEE:DICD} and LPJT (i.e., Locality Preserving Joint Transfer) \cite{IEEE:LPJT} enforced the distances of embedded representations from the same classes smaller, but the distances of embedded representations from different classes larger, thus the discriminative structure of data is maintained. This paper only focuses on discriminative preservation in both domains since it is more significant to object recognition, while the class scatter matrix is also reformulated as a probabilistic weighted manner with the proposed filtered soft labels.

%

\section{Importance Filtered Cross-Domain Adaptation}
\label{Filtered Soft Labels for Domain Adaptation}
\subsection{Preliminary}
\label{Preliminary}

\par In this paper, the bold-italic uppercase letter (i.e., $\textbf{\textit{X}}$) denotes a matrix and the bold-italic lowercase letter (i.e., $\textbf{\textit{x}}$) denotes a column vector, then $\textbf{\textit{x}}_i$ or $\textbf{\textit{X}}_{\bullet i}$ is the $i$-th column of $\textbf{\textit{X}}$ and $\textbf{\textit{X}}_{i \bullet}$ is the $i$-th row of $\textbf{\textit{X}}$. Besides, $(\textbf{\textit{X}})_{ij}$ is the value from the $i$-th row and $j$-th column of $\textbf{\textit{X}}$, and $(\textbf{\textit{x}}_i)_i$ is the $i$-th value of $\textbf{\textit{x}}_i$. 

\par The Frobenius norm of $\textbf{\textit{X}}$ is denoted by $||\textbf{\textit{X}}||_F^2$, and the $l_2$ norm of $\textbf{\textit{x}}$ is $||\textbf{\textit{x}}||_2^2$. The transpose and trace operators are denoted as $\textbf{\textit{X}}^{\top}/\textbf{\textit{x}}^{\top}$ and $tr(\textbf{\textit{X}})$. The subscript $s$ (resp. $t$) denotes the index of source domain (resp. target domain) and the superscript $c$ denotes the index of $c$-th category.

\par Therefore, the source domain (resp. target domain) could be expressed as a tri-tuple, i.e., $[\textbf{\textit{X}}_s, \textbf{\textit{y}}_s, \textbf{\textit{F}}_s]$ (resp. $[\textbf{\textit{X}}_t, \textbf{\textit{y}}_t, \textbf{\textit{F}}_t]$), where $\textbf{\textit{X}}_s\in{\mathbf{R}^{m\times n_s}}$ (resp. $\textbf{\textit{X}}_t\in{\mathbf{R}^{m\times n_t}}$), $\textbf{\textit{y}}_s\in{\mathbf{R}^{n_s}}$ (resp. $\textbf{\textit{y}}_t\in{\mathbf{R}^{n_t}}$), and $\textbf{\textit{F}}_s\in{\mathbf{R}^{(C+1)\times n_s}}$ (resp. $\textbf{\textit{F}}_t\in{\mathbf{R}^{(C+1)\times n_t}}$). Notably, all the frequently used notations are summarized in Table \ref{table_1} in this paper. 

\par In unsupervised DA, only source labels (i.e., $\textbf{\textit{y}}_s, \textbf{\textit{F}}_s$) are available but no target labels (i.e., $\textbf{\textit{y}}_t, \textbf{\textit{F}}_t$). For convenience, the tri-tuple $[\textbf{\textit{X}}, \textbf{\textit{y}}, \textbf{\textit{F}}]$ represents any given domain, and the number of samples is $n$. Notably, $\textbf{\textit{y}}$ are the digital labels, where $(\textbf{\textit{y}})_i=c$ if the $i$-th sample belongs to $c$-th class. $\textbf{\textit{F}}$ are the one-hot labels, where $argmax_j(\textbf{\textit{F}})_{ij}=c$ if the $i$-th sample pertains to $c$-th class. 

\par Moreover, given any instance $\textbf{\textit{x}}_i$ from $c$-th class, $\textbf{\textit{f}}_i$ is the hard label if only $(\textbf{\textit{f}}_i)_c=1$ but others equal to 0, while  $\textbf{\textit{f}}_i$ is the soft label if the value of each $(\textbf{\textit{f}}_i)_j$ is a confidence value between 0 and 1. Obviously, digital label must appertain to hard label since the label of each sample is entirely determined, and soft label definitely is one-hot label.      

\par This paper aims to deal with challenges in both CSDA and OSDA scenarios. Since no novel classes exist in CSDA, the values from last column of $\textbf{\textit{F}}_s$ (resp. $\textbf{\textit{F}}_t$) are 0, and the number of categories is $C$ in both $\textbf{\textit{y}}_s$ and $\textbf{\textit{y}}_t$. However, in OSDA, there are $C$ shared classes between the two domains and novel classes exist only in target domain, and those novel classes are treated as one class due to no prior knowledge about them \cite{IEEE:OSDA,IEEE:OSDA_B,IEEE:STA_OSDA}. As such, only the values from last column of $\textbf{\textit{F}}_s$ are 0, but the values from last column of $\textbf{\textit{F}}_t$ are non-zero, and the number of categories is $C$ in $\textbf{\textit{y}}_s$ but $C+1$ in $\textbf{\textit{y}}_t$.

\subsection{Revisit Cross-Domain Feature Alignment and Specific-Domain Structure Preservation}

\par In order to jointly prompt cross domain-invariant features and specific domain-invariant structures of data, it is usually better to utilize two projections to transform source and target domains data into their respective subspaces, and their new feature representations are $\textbf{\textit{Z}}_s=\textbf{\textit{A}}_s\textbf{\textit{X}}_s\in{\mathbf{R}^{k\times n_s}},\textbf{\textit{Z}}_t=\textbf{\textit{A}}_t\textbf{\textit{X}}_t\in{\mathbf{R}^{k\times n_t}}, \textbf{\textit{A}}_s,\textbf{\textit{A}}_t\in{\mathbf{R}^{m\times k}}$. Recent mainstream FDA methods have realized geometrical and statistical alignment to exploit cross domain-invariant features, and their subspace bias can be measured by Frobenius norm (i.e., $||\textbf{\textit{A}}_s-\textbf{\textit{A}}_t||_F^2$), namely subspace shift. Moreover, the MMD and class-wise MMD are used to measure their divergences in marginal and conditional distributions statistically, referred to as distributional shift, and they can be reformulated as follows
\vspace{-5pt}
\begin{equation}
\vspace{-10pt} 
||\frac{1}{n_s}\sum_{\textbf{\textit{x}}_i\in \textbf{\textit{X}}_s}\textbf{\textit{A}}_s^{\top}\textbf{\textit{x}}_i-\frac{1}{n_t}\sum_{\textbf{\textit{x}}_j\in \textbf{\textit{X}}_t}\textbf{\textit{A}}_t^{\top}\textbf{\textit{x}}_j||_2^2,
\label{eq1}
\vspace{-5pt}
\end{equation} 

\begin{equation} 
\sum_{c=1}^{C}||\frac{1}{n_s^{c}}\sum_{\textbf{\textit{x}}_i\in \textbf{\textit{X}}_s^{c}}\textbf{\textit{A}}_s^{\top}\textbf{\textit{x}}_i-\frac{1}{n_t^{c}}\sum_{\textbf{\textit{x}}_j\in \textbf{\textit{X}}_t^{c}}\textbf{\textit{A}}_t^{\top}\textbf{\textit{x}}_j||_2^2,
\label{eq2}
\vspace{-5pt}
\end{equation}

\noindent where $\textbf{\textit{X}}_s^{c}$ (resp. $\textbf{\textit{X}}_t^{c}$) is the sample subset from $c$-th class in source domain (resp. target domain), and the number of samples is $n_s^{c}$ (resp. $n_t^{c}$).  

\par To further boost specific domain-invariant discriminative structure of data, the class scatter matrix in linear discriminant analysis aims to measure inter-class and intra-class distances, and their losses are as follows
\vspace{-5pt}
\begin{equation}
\vspace{-5pt} 
tr(\sum_{c=1}^{C}n^{c}(\boldsymbol{\mu}-\boldsymbol{\mu}^{c})(\boldsymbol{\mu}-\boldsymbol{\mu}^{c})^{\top}),
\label{eq3}
\end{equation}
\begin{equation} 
tr(\sum_{c=1}^{C}\sum_{\textbf{\textit{x}}_l\in \textbf{\textit{Z}}^{c}}(\textbf{\textit{z}}_l-\boldsymbol{\mu}^{c})(\textbf{\textit{z}}_l-\boldsymbol{\mu}^{c})^{\top}),
\label{eq4}
\vspace{-5pt}
\end{equation}

\noindent where $\boldsymbol{\mu}$ is the mean of embeddings $\textbf{\textit{Z}}$, and $\boldsymbol{\mu}^{c}$ is mean of $\textbf{\textit{Z}}^{c}$, and they could be applied to either source or target domains.
 


\begin{table}[!t]
	\fontsize{6.5}{6.5}\selectfont
	\renewcommand{\arraystretch}{1.3}
	\caption{FREQUENTLY USED NOTATIONS IN THIS PAPER}
	\label{table_1}
	\centering
	\begin{tabular}{|c|c||c|c|}
		\hline
		Notation & Description & Notation & Description\\
		\hline
		\hline
		$m$ & original dimensions & $\textbf{\textit{X}}_s/\textbf{\textit{X}}_t$ & source/target data\\
		\hline
		$n_s/n_t/n_{st}$ & source/target/all samples & $\textbf{\textit{y}}_s/\textbf{\textit{y}}_t$ & source/target hard labels\\
		\hline
		$C+1$ & shared+novel classes & $\textbf{\textit{F}}_s/\textbf{\textit{F}}_t$ & source/target soft labels\\
		\hline
		$k$ & projected dimensions & $\textbf{\textit{A}}_s/\textbf{\textit{A}}_t$ & source/target projection\\
		\hline
		$\boldsymbol{\mu}$ & mean vector & $\textbf{\textit{Z}}_s/\textbf{\textit{Z}}_t$ & source/target embeddings\\
		\hline
		$N$ & selected number & $\textbf{\textit{P}}$ & joint projection\\
		\hline
		$p$ & neighbors number & $\textbf{\textit{W}}$ & similarity matrix\\
		\hline
		$\alpha_{set},\gamma,\beta$ & regularization parameters & $\textbf{\textit{M}}_1,\textbf{\textit{M}}_2$ & MMD matrix\\
		\hline
		$\lambda,\delta$ & regularization parameters & $\textbf{\textit{N}}_{b},\textbf{\textit{N}}_{w}$ & class scatter matrix\\
		\hline
	\end{tabular}
	\vspace{-10pt}
\end{table}

\subsection{Motivation}
\label{Motivation}

\par Actually, a majority of FDA methods employed digital hard labels to iteratively compute Eq. \ref{eq2} $\sim$ Eq. \ref{eq4}, where target labels are previously predicted using some base classifiers namely pseudo labels \cite{IEEE:JDA}. Since target labels are pseudo, it is overconfident to utilize their hard expressions at first hand, and easily to cause negative transfer. Intuitively, soft labels could compensate for the drawback of hard ones since confidence values between 0 and 1 are allocated to each class probability. 

\par Insightfully, we observed that soft labels are so confusing that negative transfer still remains, due to hard target samples existed in real life, e.g., those samples distributed in class overlaps. Therefore, this paper proposes an importance filtered mechanism to fall away small noisy probabilities but retain those prominent ones in soft labels, thus their adverse effect on knowledge transfer could be mitigated greatly. 

\par While the filtering principle is contrived, some difficulties still exist. Firstly, a general classifier have to be well-designed, thus it could not only output soft labels, but also be capable of detecting novel classes in OSDA scenario. Once the filtered soft labels are obtained, then we have to utilize such one-hot labels to devise general expressions of Eq. \ref{eq1} $\sim$ Eq. \ref{eq4}, such that they could address the challenges of CSDA and OSDA simultaneously. Specially, Eq. \ref{eq1} should be reformulated as a category-relevant form like class-wise MMD.

\subsection{Importance Filtered Mechanism}
\label{Filtered Soft Labels}

\begin{figure}[!t]
	\centering
	\includegraphics[width=1.0\linewidth,height=0.16\textheight]{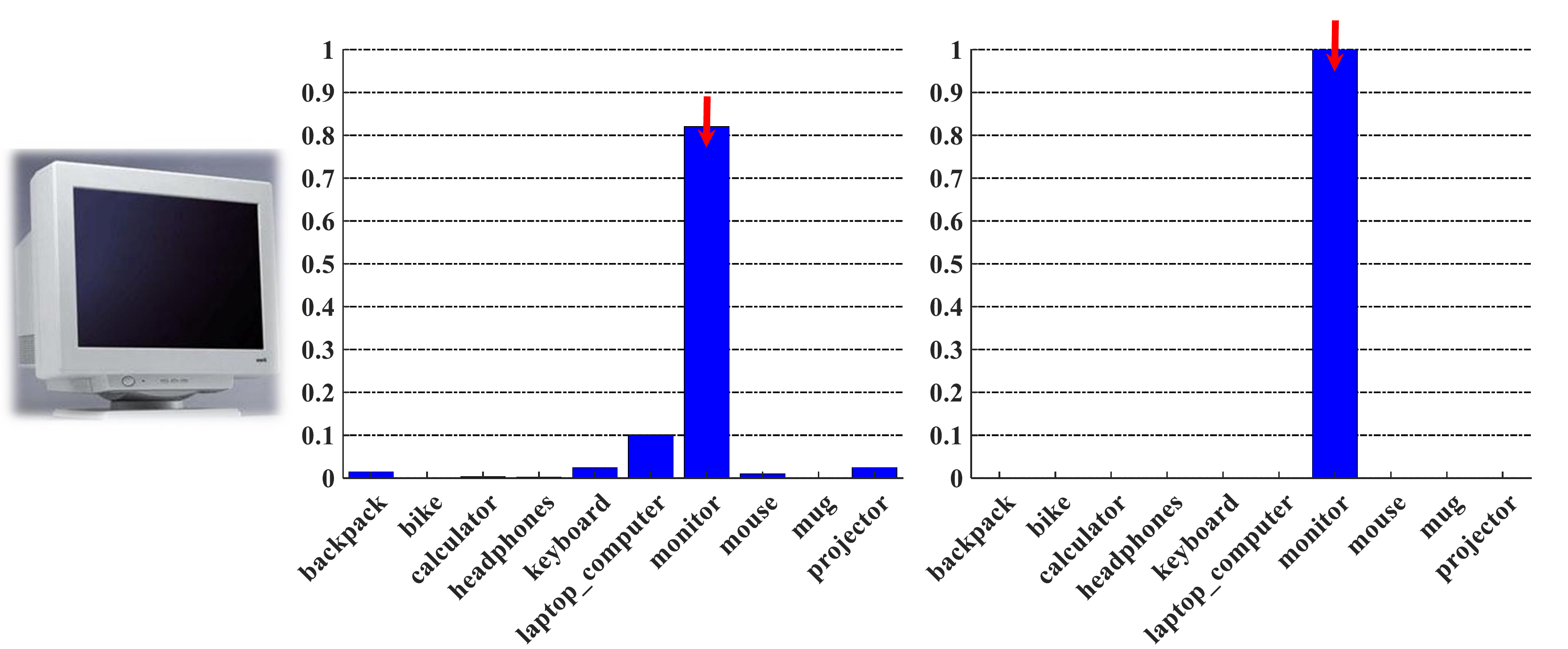}
	\vspace{-20pt}
	\caption{Importance filtered confident labels.}
	\vspace{-15pt}
	\label{fig1}
\end{figure}

\begin{figure}[!t]
	\centering
	\includegraphics[width=1.0\linewidth,height=0.16\textheight]{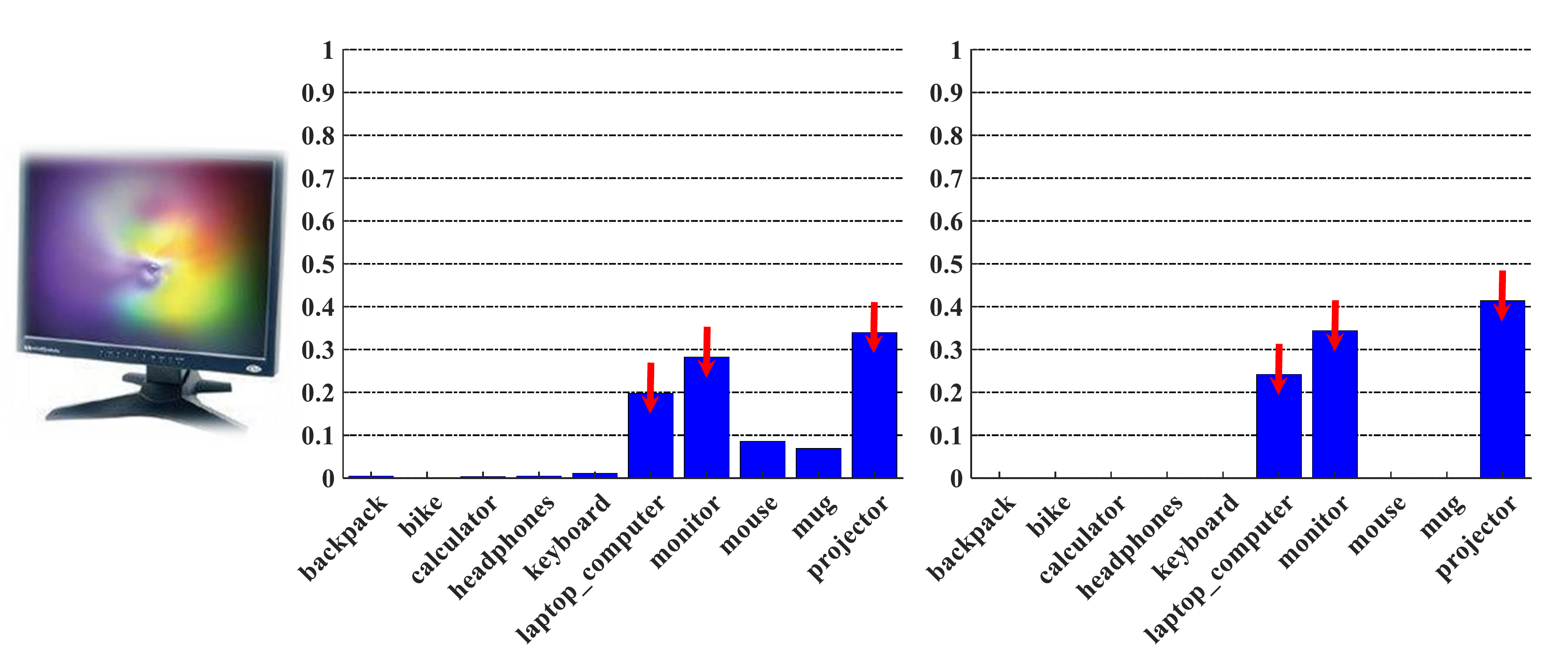}
	\vspace{-20pt}
	\caption{Importance filtered ambiguous labels.}
	\vspace{-15pt}
	\label{fig2}
\end{figure}

\par Intriguingly, we propose a simple but highly effective importance filtered mechanism. Since the source labels are available, only target soft labels are in this filtered progress. Specifically, the target soft labels could be broken down into two types:

\noindent 1) the confident labels whose maximum probabilistic values in their label vectors exceed a given threshold, which is empirically set to 0.8 in this paper. 

\noindent 2) the ambiguous labels whose maximum probabilistic values in their label vectors are no more than 0.8. 
\par With regard to the confident labels, we directly obtain their hard one-hot labels by only retaining the maximum probabilities. Concerning the ambiguous labels, we select several representative probabilistic values by setting a hyper-parameter $N$, but the remaining values are set to 0. Then we normalize each advanced label vector so that its sum equals to 1. To clearly express the proposed importance filtered mechanism, we illustrate these two cases in Fig. \ref{fig1} and Fig. \ref{fig2} when $N=3$.


\par To summarize, this importance filtered mechanism could throw away negligible noisy probabilistic values in those two kinds of soft labels, while nearly all of the prominent ones are retained desirably. As such, negative transfer incurred by both hard and soft labels could be mitigated significantly.

\subsection{A General Graph-based Label Propagation}
\label{A General Graph-based Label Propagatio}

\par As a label prediction algorithm, Graph-based Label Propagation (GLP) aims to propagate label information from labeled data (i.e., source domain) to unlabeled data (i.e., target domain) by establishing a neighborhood similarity graph on the whole data. We adopt a popular graph construction method as follows: if $\textbf{\textit{x}}_i$ is among the $p$-nearest neighbors of $\textbf{\textit{x}}_j$ or $\textbf{\textit{x}}_j$ is among the $p$-nearest neighbors of $\textbf{\textit{x}}_i$, then $\textbf{\textit{x}}_i$ and $\textbf{\textit{x}}_j$ are linked by a similarity weight computed by $(\textbf{\textit{W}})_{ij}=e^{-||\textbf{\textit{x}}_i-\textbf{\textit{x}}_j||_2^2/\sigma^2}$, otherwise, $(\textbf{\textit{W}})_{ij}=0$, where $\sigma$ is the variance. To deal with both CSDA and OSDA scenarios, we present a General Graph-based Label Propagation (GGLP) method as follows
\begin{equation} 
\min\limits_{\textbf{\textit{F}}^*}\sum_{i,j=1}^{n_{st}}(\textbf{\textit{W}})_{ij}||\textbf{\textit{f}}_i^*-\textbf{\textit{f}}_j^*||_2^2+\sum_{l=1}^{n_{st}}u_lh_l||\textbf{\textit{f}}_l^*-\textbf{\textit{f}}_l||_2^2,
\label{eq5}
\vspace{-5pt}
\end{equation}

\noindent where $u_l$ is a regularization parameter for each data $\textbf{\textit{x}}_l$, and $\textbf{\textit{f}}_l\in\textbf{\textit{F}}=[\textbf{\textit{F}}_s,\textbf{\textit{F}}_t]$, $h_l=\sum_{i}(\textbf{\textit{W}})_{il}$. In CSDA setting, $\textbf{\textit{F}}_t$ is initialized as zero matrix due to no target labels, and $\textbf{\textit{F}}^*$ is optimization variable. Let us introduce a set of variables $\alpha_l=1/(1+u_l), (l=1,...,n_{st})$. The label information of each data is partly received from its neighbors’ labels, and the rest is received from its initial label
$\textbf{\textit{f}}_l$. For a data instance from source domain, we are sure that its initial label is correct confidently, $\alpha_l$ can be set to 0, i.e., $u_l\rightarrow +\infty$, which means that the resulted label $\textbf{\textit{f}}_l^*$ of $\textbf{\textit{x}}_l$ will equal to the initial label $\textbf{\textit{f}}_l$ and remain unchanged. Thus the classification errors in source domain could be reduced 0. For the data point from target domain, $\alpha_l$ can be set to one, since the
label of $\textbf{\textit{x}}_l$ is unknown and its label is determined by its neighbor's.

\par In OSDA setting, $(\textbf{\textit{F}}_t)_{ij}=1$ if $i=C+1$ and $(\textbf{\textit{F}}_t)_{ij}=0$ otherwise. Similarly, we set $\alpha_l$ as 0 in the source domain to enable their resulted labels unchanged. However, $\alpha_l$ should be set as a positive value between one and zero in the target domain, and the smaller $\alpha_l$ is, the easier one sample is to be classified as novel classes. For example, the extreme case $\alpha_l=1$, i.e., $u_l=0$, which means that the resulted label $\textbf{\textit{f}}_l^*$ of $\textbf{\textit{x}}_l$ could deviate from initial label $\textbf{\textit{f}}_l^*$ to the most extent. Since the last component of $\textbf{\textit{f}}_l$ in target domain is initialized as 1, i.e., definitely pertains to novel classes, it would lose the capability to discover novel classes when $\alpha_l=1$. Another extreme case is $\alpha_l=0$, i.e., $u_l\rightarrow +\infty$, which means that the resulted label $\textbf{\textit{f}}_l^*$ of $\textbf{\textit{x}}_l$, will be confidently classified as novel class, and thus lose the identify ability. In the next two sections, we will employ the proposed one-hot filtered labels to formulate the general expressions of Eq. \ref{eq1} $\sim$ Eq. \ref{eq4}.  

\subsection{Filtered Soft Labels-based distributional shift}
\label{Filtered Soft Labels-based distributional shift}

\par As regards the MMD term, we reformulate it as follows
\vspace{-5pt}
\begin{equation} 
||\frac{1}{\widehat{n}_s^{1}}\sum_{i=1}^{n_s}(\widehat{\textbf{\textit{F}}}_{s})_{1i}\textbf{\textit{A}}_s^{\top}\textbf{\textit{x}}_i-\frac{1}{\widehat{n}_t^{1}}\sum_{j=1}^{n_t}(\widehat{\textbf{\textit{F}}}_{t})_{1j}\textbf{\textit{A}}_t^{\top}\textbf{\textit{x}}_j||_2^2,
\label{eq6}
\vspace{-5pt}
\end{equation}

\noindent where $\widehat{n}_s^{1}=\sum_{i=1}^{n_s}(\widehat{\textbf{\textit{f}}}_{s}^1)_i$ and $\widehat{n}_t^{1}=\sum_{j=1}^{n_t}(\widehat{\textbf{\textit{f}}}_{t}^1)_j$. In OSDA setting, our goal is to align distributions of shared classes from source and target domains. The source and target labels could be divided into two categories (i.e., shared and novel classes). Hence, we denote the newly importance filtered soft labels as $\widehat{\textbf{\textit{F}}}_s=[\widehat{\textbf{\textit{f}}}_{s}^1;\widehat{\textbf{\textit{f}}}_{s}^2]$ and $\widehat{\textbf{\textit{F}}}_t=[\widehat{\textbf{\textit{f}}}_{t}^1;\widehat{\textbf{\textit{f}}}_{t}^2]$, where $\widehat{\textbf{\textit{f}}}_{s}^1=\sum_{c=1}^{C}(\textbf{\textit{F}}_s)_{c\bullet}$ (resp. $\widehat{\textbf{\textit{f}}}_{t}^1=\sum_{c=1}^{C}(\textbf{\textit{F}}_t)_{c\bullet}$) and $\widehat{\textbf{\textit{f}}}_{s}^2=(\textbf{\textit{F}}_s)_{(C+1)\bullet}$ (resp. $\widehat{\textbf{\textit{f}}}_{t}^2=(\textbf{\textit{F}}_t)_{(C+1)\bullet}$). It is worth noting that Eq. \ref{eq1} is a special case of Eq. \ref{eq6}, when there are no novel classes in target domain (i.e., CSDA setting). For convenience of matrix calculus, Eq. \ref{eq6} can be rewritten as $Tr(\textbf{\textit{P}}^{\top}\textbf{\textit{M}}_1\textbf{\textit{P}})$, where $\textbf{\textit{P}}=[\textbf{\textit{A}}_S;\textbf{\textit{A}}_T]$ is the joint projection matrix, and $\textbf{\textit{M}}_1=[\textbf{\textit{M}}_{ss1}^1,\textbf{\textit{M}}_{st1}^1;\textbf{\textit{M}}_{ts1}^1,\textbf{\textit{M}}_{tt1}^1]$. $\textbf{\textit{M}}_{ss1}^1$, $\textbf{\textit{M}}_{tt1}^1$, $\textbf{\textit{M}}_{st1}^1$, $\textbf{\textit{M}}_{ts1}^1$ are computed as follows
\begin{equation} 
\vspace{-5pt}
\begin{array}{lr}
\textbf{\textit{M}}_{ss1}^1=\textbf{\textit{X}}_s\textbf{\textit{Q}}_{ss1}^1\textbf{\textit{X}}_{s}^{\top},\quad
\textbf{\textit{Q}}_{ss1}^1=\widehat{\textbf{\textit{F}}}_s^{1\ast}\textbf{\textit{e}}_s^1{\textbf{\textit{e}}_s^1}^{\top} \widehat{\textbf{\textit{F}}}_s^{1\ast},
\\
\\
\widehat{\textbf{\textit{F}}}_s^{1\ast}=diag(\widehat{\textbf{\textit{f}}}_s^1), \quad \textbf{\textit{e}}_s^1=\textbf{\textit{1}}_{ns}/\widehat{n}_s^{1},
\\
\\
\textbf{\textit{M}}_{tt1}^1=\textbf{\textit{X}}_t\textbf{\textit{Q}}_{tt1}^1\textbf{\textit{X}}_{t}^{\top},\quad
\textbf{\textit{Q}}_{tt1}^1=\widehat{\textbf{\textit{F}}}_t^{1\ast}\textbf{\textit{e}}_t^1{\textbf{\textit{e}}_t^1}^{\top} \widehat{\textbf{\textit{F}}}_t^{1\ast},
\\
\\
\widehat{\textbf{\textit{F}}}_t^{1\ast}=diag(\widehat{\textbf{\textit{f}}}_t^1), \quad \textbf{\textit{e}}_t^1=\textbf{\textit{1}}_{nt}/\widehat{n}_t^{1},
\\
\\
\textbf{\textit{M}}_{st1}^1=\textbf{\textit{X}}_s\textbf{\textit{Q}}_{st1}^1\textbf{\textit{X}}_{t}^{\top},\quad
\textbf{\textit{Q}}_{st1}^1=\widehat{\textbf{\textit{F}}}_s^{1\ast}\textbf{\textit{e}}_s^1{\textbf{\textit{e}}_t^1}^{\top} \widehat{\textbf{\textit{F}}}_t^{1\ast},
\\
\\
\textbf{\textit{M}}_{ts1}^1=\textbf{\textit{X}}_t\textbf{\textit{Q}}_{tt1}^1\textbf{\textit{X}}_{s}^{\top},\quad
\textbf{\textit{Q}}_{ts1}^1=\widehat{\textbf{\textit{F}}}_t^{1\ast}\textbf{\textit{e}}_t^1{\textbf{\textit{e}}_s^1}^{\top} \widehat{\textbf{\textit{F}}}_s^{1\ast}.
\end{array}
\label{eq7}
\end{equation}
\vspace{5pt}
\par Similar to MMD, class-wise MMD can be reformulated as follows
\vspace{-5pt}
\begin{equation} 
\sum_{c=1}^{C}||\frac{1}{\widehat{n}_s^{c}}\sum_{i=1}^{n_s}(\textbf{\textit{F}}_s)_{ci}\textbf{\textit{A}}_s^{\top}\textbf{\textit{x}}_i-\frac{1}{\widehat{n}_t^{c}}\sum_{j=1}^{n_t}(\textbf{\textit{F}}_t)_{cj}\textbf{\textit{A}}_t^{\top}\textbf{\textit{x}}_j||_2^2.
\label{eq8}
\end{equation}

\noindent It can be noticed that Eq. \ref{eq6} is a special case of Eq. \ref{eq8} when $C=1$, thus Eq. \ref{eq8} equals to $Tr(\textbf{\textit{P}}^{\top}\textbf{\textit{M}}_2\textbf{\textit{P}})$, and $\textbf{\textit{M}}_2=\sum_{c=1}^{C}[\textbf{\textit{M}}_{ss2}^c,\textbf{\textit{M}}_{st2}^c;\textbf{\textit{M}}_{ts2}^c,\textbf{\textit{M}}_{tt2}^c]$. $\textbf{\textit{M}}_{ss2}^c$, $\textbf{\textit{M}}_{tt2}^c$, $\textbf{\textit{M}}_{st2}^c$, $\textbf{\textit{M}}_{ts2}^c$ can be computed by Eq. \ref{eq7} with corresponding filtered soft labels.

\subsection{Filtered Soft Labels-based Discriminant Analysis}
\label{Filtered Soft Labels-based Discriminant Analysis}

\par We can do the similar reconstruction for class scatter matrix in linear discriminant analysis, and the formulas are as follows
\begin{equation} 
\begin{array}{lr}
tr(\sum_{c=1}^{C}\frac{\widehat{n}^{c}}{\widehat{n}}(\widehat{\boldsymbol{\mu}}-\widehat{\boldsymbol{\mu}}^{c})(\widehat{\boldsymbol{\mu}}-\widehat{\boldsymbol{\mu}}^{c})^{\top})=Tr(\textbf{\textit{P}}^{\top}\textbf{\textit{N}}_b\textbf{\textit{P}})
\\
\\
=tr(\textbf{\textit{P}}^{\top}\frac{1}{\widehat{n}}\textbf{\textit{X}}(\textbf{\textit{F}}^{\top}\textbf{\textit{K}}\textbf{\textit{F}}-\frac{1}{\widehat{n}}\textbf{\textit{B}}\textbf{\textit{1}}_n\textbf{\textit{1}}_n^{\top}\textbf{\textit{B}})\textbf{\textit{X}}^{\top}\textbf{\textit{P}}),
\end{array}
\label{eq9}
\end{equation}
\begin{equation} 
\begin{array}{lr}
\sum_{c=1}^{C}\sum_{\textbf{\textit{z}}_l\in \textbf{\textit{Z}}^{c}}\frac{1}{\widetilde{n}}\textbf{\textit{f}}_{cl}(\textbf{\textit{z}}_l-\widetilde{\boldsymbol{\mu}}^{c})(\textbf{\textit{z}}_l-\widetilde{\boldsymbol{\mu}}^{c})^{\top}=Tr(\textbf{\textit{P}}^{\top}\textbf{\textit{N}}_w\textbf{\textit{P}}),
\\
\\
=tr(\textbf{\textit{P}}^{\top}\frac{1}{\widehat{n}}\textbf{\textit{X}}(\textbf{\textit{B}}-\textbf{\textit{F}}^{\top}\textbf{\textit{K}}\textbf{\textit{F}})\textbf{\textit{X}}^{\top}\textbf{\textit{P}}),
\end{array}
\label{eq10}
\end{equation}

\noindent where $\widehat{n}^{c}=\sum_{l}\textbf{\textit{f}}_{cl}$ and $\widehat{n}=\sum_{c}\widehat{n}^{c}$, and $\widehat{\boldsymbol{\mu}}^{c}=\sum_{l}\textbf{\textit{f}}_{cl}\textbf{\textit{z}}_l/\widehat{n}^{c}$ and $\widehat{\boldsymbol{\mu}}=\sum_{c}\sum_{l}\textbf{\textit{f}}_{cl}\textbf{\textit{z}}_l/\widehat{n}$. $\textbf{\textit{N}}_b,\textbf{\textit{N}}_w\in \mathbf{R}^{m\times m}$ are the class scatter matrix. $\textbf{\textit{B}}\in \mathbf{R}^{n\times n}$ and $\textbf{\textit{K}}\in \mathbf{R}^{C\times C}$ are diagonal matrix, where $(\textbf{\textit{B}})_{ii}=\sum_{j=1}^{C}\textbf{\textit{F}}_{ji}$ and $(\textbf{\textit{K}})_{ii}=1/\sum_{j=1}^{n}\textbf{\textit{F}}_{ij}$, and $\textbf{\textit{1}}_n\in\mathbf{R}^{n\times 1}$ is a column vector whose elements are all 1. 

\subsection{Proposed Approach}  
\label{Proposed Approach}
\par After those significant losses in DA are re-established, the proposed approach could be formulated as follows
\begin{equation} 
\min_{\textbf{\textit{F}}^*}tr(\textbf{\textit{F}}^{*\top}\textbf{\textit{L}}\textbf{\textit{F}}^*)+tr((\textbf{\textit{F}}^*-\textbf{\textit{F}})\textbf{\textit{U}}\textbf{\textit{H}}(\textbf{\textit{F}}^*-\textbf{\textit{F}})^{\top}),
\label{eq11}
\end{equation}
\vspace{-10pt}
\begin{equation} 
\max_{\textbf{\textit{P}}}
\frac{Tr(\textbf{\textit{P}}^{\top}[\textbf{\textit{N}}_{sb},\textbf{\textit{0}};\textbf{\textit{N}}_{tb},\textbf{\textit{0}}]\textbf{\textit{P}})}{Tr(\textbf{\textit{P}}^{\top}(\lambda[\textbf{\textit{N}}_{sw},\textbf{\textit{0}};\textbf{\textit{N}}_{tw},\textbf{\textit{0}}]+\delta(\textbf{\textit{M}}_1+\textbf{\textit{M}}_2)+\textbf{\textit{V}})\textbf{\textit{P}})},
\label{eq12}
\end{equation}

\noindent where $\textbf{\textit{U}}, \textbf{\textit{H}}\in \mathbf{R}^{n_{st}\times n_{st}}$ are diagonal matrix, $(\textbf{\textit{U}})_{ii}=u_i$, $(\textbf{\textit{H}})_{ii}=h_i$, and $\textbf{\textit{L}}=\textbf{\textit{H}}-\textbf{\textit{W}}$. Moreover, $Tr(\textbf{\textit{P}}^{\top}\textbf{\textit{V}}\textbf{\textit{P}})=\beta||\textbf{\textit{A}}_s-\textbf{\textit{A}}_t||_F^2+\gamma(||\textbf{\textit{A}}_s||_F^2+||\textbf{\textit{A}}_t||_F^2)$ is the subspace shift loss and constraint on scale of the two projections as \cite{IEEE:DICD,IEEE:JGSA} did, where $\textbf{\textit{V}}=[\textbf{\textit{I}}_{(\beta+\gamma)},-\textbf{\textit{I}}_{\beta};-\textbf{\textit{I}}_{\beta},\textbf{\textit{I}}_{(\beta+\gamma)}]$, $\textbf{\textit{I}}$ is an identify matrix and $\textbf{\textit{I}}_{(\beta+\gamma)/\beta}$ is a diagonal matrix with $i$-th diagonal entry being $(\beta+\gamma)/\beta$. Eq. \ref{eq11} is the procedure of label prediction, and this well-designed GGLP classifier is capable of not only feeding filtered soft labels to Eq. \ref{eq12}, but also detecting novel classes in OSDA scenario. 

\par Eq. \ref{eq12} is the process of transferable features learning, the subspace alignment is realized by minimizing the Frobenius norm between their respective subspaces. The cross domain-invariant features are leveraged by decreasing marginal and conditional distribution differences simultaneously, and the specific domain-invariant discriminative structure of data in both domains is promoted by maximizing the inter-class dispersion but minimizing the intra-class scatter. Differently, filtered soft labels are used to reformulate those losses, thus negative transfer can be mitigated greatly incurred by hard and soft labels. Moreover, the proposed general losses could deal with both CSDA and OSDA settings. Remarkably, by alternatively optimize Eq. \ref{eq11} and Eq. \ref{eq12}, the neighborhood similarity graph in Eq. \ref{eq11} established with transferable features learned from Eq. \ref{eq12}, could prompt more accurate soft labels. In return, when more accurate soft labels in Eq. \ref{eq11} are refined using proposed filtering mechanism, more positive knowledge transfer could be obtained in Eq. \ref{eq12} by reformulating those significant losses in DA.    
   
\subsection{Solution to Our Proposed Model}
\label{Optimization}
\noindent \textbf{Label Prediction:}

\par We utilize the transferred features $\textbf{\textit{Z}}_s, \textbf{\textit{Z}}_t$ during the current iteration to compute $\textbf{\textit{L}}$, thus we obtain the partial derivative of Eq. \ref{eq11} w.r.t., $\textbf{\textit{F}}^*$, by setting it to $\textbf{\textit{0}}$ as
\begin{equation} 
\textbf{\textit{L}}\textbf{\textit{F}}^*+\textbf{\textit{U}}\textbf{\textit{H}}(\textbf{\textit{F}}^*-\textbf{\textit{F}})=\textbf{\textit{0}},
\label{eq13}
\end{equation}
\noindent then the solution can be derived as follows
\begin{equation} 
\textbf{\textit{F}}^*=(\textbf{\textit{I}}-\textbf{\textit{I}}_{\alpha}\textbf{\textit{H}}^{-1}\textbf{\textit{W}})(\textbf{\textit{I}}-\textbf{\textit{I}}_{\alpha})\textbf{\textit{F}}.
\label{eq14}
\end{equation}
\noindent \textbf{Projection Learning:}
\par Similar to previous work \cite{IEEE:DICD,IEEE:JGSA}, Eq. \ref{eq12} could be rewritten as follows

\begin{equation} 
\begin{array}{lr}
\qquad \qquad \max\limits_{\textbf{\textit{P}}}\quad Tr(\textbf{\textit{P}}^{\top}[\textbf{\textit{N}}_{sb},\textbf{\textit{0}};\textbf{\textit{N}}_{tb},\textbf{\textit{0}}]\textbf{\textit{P}}) \\
s.t. \quad Tr(\textbf{\textit{P}}^{\top}(\lambda[\textbf{\textit{N}}_{sw},\textbf{\textit{0}};\textbf{\textit{N}}_{tw},\textbf{\textit{0}}]+\delta(\textbf{\textit{M}}_1+\textbf{\textit{M}}_2)+\textbf{\textit{V}})\textbf{\textit{P}}).
\end{array}
\label{eq15}
\end{equation}

\noindent According to the constrained optimization theory, we denote $\boldsymbol{\Phi}=diag(\phi_1,...,\phi_k)\in{\mathbf{R}^{k\times{k}}}$ as the Lagrange multiplier, and derive the Lagrange function for Eq. \ref{eq15}, w.r.t., $\textbf{\textit{P}}$ as follows
\begin{equation}
[\textbf{\textit{N}}_{sb},\textbf{\textit{0}};\textbf{\textit{N}}_{tb},\textbf{\textit{0}}]\textbf{\textit{P}}=(\lambda[\textbf{\textit{N}}_{sw},\textbf{\textit{0}};\textbf{\textit{N}}_{tw},\textbf{\textit{0}}]+\delta(\textbf{\textit{M}}_1+\textbf{\textit{M}}_2)+\textbf{\textit{V}})\textbf{\textit{P}}\boldsymbol{\Phi}.
\label{eq16}
\end{equation}
\noindent Finally, finding the solution of $\textbf{\textit{P}}$ is reduced
to solve Eq. \ref{eq16} for the $k$ leading eigenvalues, and the corresponding eigenvectors, which can be solved analytically through generalized eigenvalue decomposition. Once the joint projection $\textbf{\textit{P}}$ is obtained, their respective projections $\textbf{\textit{A}}_s$ and $\textbf{\textit{A}}_t$ can be obtained easily.

\par Notably, by alternating those two steps, we will iteratively enhance the quality of predicted labels and transferred features, and a complete procedure of IFCDA for challenges of both CSDA and OSDA scenarios is summarized in Algorithm \ref{alg1}.

\begin{algorithm}
	\caption{Importance Filtered Cross-Domain Adaptation}
	\label{alg1}
\begin{algorithmic}[1]
\renewcommand{\algorithmicrequire}{\textbf{Input:}}
\REQUIRE Labeled source domain $[\textbf{\textit{X}}_s, \textbf{\textit{F}}_s]$, Unlabeled target domain $\textbf{\textit{X}}_t$, parameters $p$, $N$, $\alpha_{set}$, $T$, $k$, $\gamma$, $\beta$, $\lambda$, $\delta$
\STATE $\textbf{\textit{L}}$ computed by $[\textbf{\textit{X}}_s, \textbf{\textit{X}}_t]$ 
\IF{CSDA scenario}
\STATE $\alpha_i=0$ for source data but $\alpha_i=1$ for target data 
\STATE $\textbf{\textit{F}}_t=\textbf{\textit{0}}_{(C+1)\times n_t}$
\STATE Obtain $\textbf{\textit{F}}_t^*$ using Eq. \ref{eq14} and filter soft labels $\textbf{\textit{F}}_t^*$ using $N$ as Section \ref{Filtered Soft Labels} does 
\ELSE
\STATE OSDA scenario \textbf{then}
\STATE $\alpha_i=0$ for source data but $\alpha_i=\alpha_{set}$ for target data
\STATE $\textbf{\textit{F}}_t=[\textbf{\textit{0}}_{C\times n_t};\textbf{\textit{0}}_{1\times n_t}]$
\STATE Obtain $\textbf{\textit{F}}_t^*$ using Eq. \ref{eq14} and filter soft labels $\textbf{\textit{F}}_t^*$ using $N$ as Section \ref{Filtered Soft Labels} does 
\ENDIF
\WHILE{$iter \leq T$}
\STATE Compute $\textbf{\textit{M}}_1$, $\textbf{\textit{M}}_2$, $\textbf{\textit{N}}_{sb}$, $\textbf{\textit{N}}_{sw}$ as Section \ref{Filtered Soft Labels-based distributional shift} and Section \ref{Filtered Soft Labels-based Discriminant Analysis} do
\STATE Obtain $\textbf{\textit{P}}=[\textbf{\textit{A}}_s;\textbf{\textit{A}}_t]$ using Eq. \ref{eq16}
\STATE Obtain $\textbf{\textit{Z}}=[\textbf{\textit{Z}}_s,\textbf{\textit{Z}}_t], \textbf{\textit{Z}}_s=\textbf{\textit{A}}_s\textbf{\textit{X}}_s, \textbf{\textit{Z}}_t=\textbf{\textit{A}}_t\textbf{\textit{X}}_t$
\STATE $\textbf{\textit{L}}$ computed by $[\textbf{\textit{Z}}_s, \textbf{\textit{Z}}_t]$ 
\STATE Conduct line 2 $\sim$ line 10
\ENDWHILE
\renewcommand{\algorithmicrequire}{\textbf{Output:}}
\REQUIRE Target labels $\textbf{\textit{F}}_t^*$
\end{algorithmic}
\end{algorithm}
\subsection{Computational Complexity}
\label{Computational Complexity}
\par Now we analyze the computational complexity of Algorithm \ref{alg1} by the big $O$ notation. The time cost of Algorithm \ref{alg1} consists of the following two parts:
\\
1) Computing $\textbf{\textit{M}}_1$, $\textbf{\textit{M}}_2$, $\textbf{\textit{N}}_{sb}$, $\textbf{\textit{N}}_{sw}$ in line 13 costs $O(Tmn_t^2+Tmn_t^2+TCn_t^2+TCn_t^2+TCn_s^3+TCn_t^3+TCn_sn_t^2+TCn_s^2n_t)$.
\\
2) Solving Eq. \ref{eq15} in line 14 costs $O(Tmk^2)$.

\par Then, the overall computational complexity of Algorithm \ref{alg1} is $O(Tmn_t^2+Tmn_t^2+TCn_t^2+TCn_t^2+TCn_s^3+TCn_t^3+TCn_sn_t^2+TCn_s^2n_t+Tmk^2)$. In section \ref{Model Analysis}, we will show that the number of iterations $T$ is usually smaller than 10, which is enough to guarantee the convergence. Besides, the typical values of $k$ are not greater than 200, so $T,k\ll{min(m,n)}$. Therefore, Algorithm \ref{alg1} can be solved in polynomial time with respect to the number of samples.

\section{Experiments} 
\label{Experiments}
\subsection{Datasets Description}
\label{Datasets Description}
\par We employed the real datasets, i.e., USPS, MNIST, COIL-20, MSRC, VOC-2007, Office-31, Caltech-256, Office-Home, which is widely applied in cross-domain classification problem, to evaluate the effectiveness of the proposed approach either in CSDA or OSDA settings. By combining them with different arranges, this paper established 6 CSDA datasets (i.e., USPS vs MNIST, COIL-1 vs COIL-2, MSRC vs VOC-2007, Office-10 vs Caltech-10, Office-Home), and 2 OSDA datasets (i.e., Office-31, Office-Home). The datasets description is given as follows.

\subsubsection{USPS vs MNIST}
By combining USPS and MNIST, the CSDA dataset USPS vs MNIST is created, and they follow very different distributions due to various writing styles and illustration intensity. USPS consists of 9,298 images of size 16$\times$16 and 70,000 images of size 28$\times$28 in MNIST, and 10 categories are shared across them. Followed by previous work \cite{IEEE:JDA,IEEE:JGSA}, we adopted their subsets where 1,800 examples are randomly sampled from USPS and 2,000 instances from MNIST. For convenience, we denote USPS and MNIST as U and M, then 2 CSDA tasks can be constructed, i.e., U$\rightarrow$M, M$\rightarrow$U. Note that the arrow "$\rightarrow$" in this paper is the direction from source to target. For example, U$\rightarrow$M means that U is the labeled source domain and M is the unlabeled target domain. Additionally, all images are rescaled uniformly to size 16$\times$16, and each sample image is encoded by a 256-dimension feature vector of gray-scale pixel values.     

\subsubsection{COIL-1 vs COIL-2}
COIL-1 (C-1) and COIL-2 (C-2) are sampled from COIL-20 and they follow different distributions, since C-1 locates in $[0^\circ,85^\circ]\cup[180^\circ,265^\circ]$ but C-2 locates in $[90^\circ,175^\circ]\cup[270^\circ,355^\circ]$. COIL-20 is sampled from 20 objects of size 32$\times$32 and consists of 1,440 images, where each object rotates 5 degrees successively and derives 72 images. Likewise, 2 CSDA tasks can be established, i.e., C-1$\rightarrow$C-2, C-2$\rightarrow$C-1, and each sample image is encoded by a 1024-dimension feature vector of gray-scale pixel values.  

\subsubsection{MSRC vs VOC-2007}
MSRC (Ms) and VOC-2007 (Vo) are also originated from very different distributions, because Ms is the standard image dataset, but Vo is the digital photo dataset in Flickr-7. Ms has more than 4,000 samples from 18 categories, and Vo contains over 5,000 samples annotated with 20 concepts. Followed by previous work \cite{IEEE:TSC,IEEE:TRSC,IEEE:RTML}, we selected 6 shared semantic classes where 1,269 images in Ms and 1,530 images in Vo. Similarly, 2 CSDA tasks can be erected, i.e., Ms$\rightarrow$Vo, Vo$\rightarrow$Ms, and Dense SIFT features with 128 dimensions are used in this paper.

\subsubsection{Office-31}
Office-31 consists of 3 domains, i.e., Amazon (A), Webcam (W), DSLR (D), and contains 4,652 images from 31 categories, they follow different distributions to each other. We did the same setting as previous OSDA work \cite{IEEE:STA_OSDA,IEEE:OSDA,IEEE:OSDA_B}, i.e, the same set of known shared classes across the two domains and private unknown classes in the target domain. As such, 3$\times$2=6 OSDA tasks could be created. For a fair comparison, the deep features are adopted, which is pre-extracted from the AlexNet model (i.e., Fc-7 features with 4096 dimensions) \cite{IEEE:AlexNet} and ResNet-50 model (i.e., ResNet-50 features with 2048 dimensions) \cite{IEEE:ResNet}.  

\subsubsection{Office-10 vs Caltech-10}
Caltech-256 (C) involves over 30,000 samples from 256 categories, which obeys very different distribution compared with A, W, D in Office-31. Followed by previous work \cite{IEEE:JDA,IEEE:JGSA,IEEE:GFK}, 10 shared categories were deliberately picked from C, A, W, D, i.e., the CSDA dataset Office-10 vs Caltech-10. Therefore, 4$\times$3=12 CSDA tasks could be established, and the SURF features with 800 dimensions are utilized here.

\subsubsection{Office-Home}
Office-Home \cite{IEEE:DHN} is a more challenging dataset, which has been released recently and crawled through several search engines and online image directories. It consists of 4 different domains, i.e., Artistic images (Ar), Clipart images (Cl), Product images (Pr) and Real-World images (Rw), and they follow very different distributions to each other. In total, there are 65 object categories for each domain and 15,500 images in the whole dataset. Concerning CSDA setting, we utilized the images from all 65 categories, and constructed 4$\times$3=12 CSDA tasks. Similar to previous OSDA work \cite{IEEE:STA_OSDA,IEEE:OSDA_B}, we chose the first 25 classes (i.e., in alphabetic order) shared by the source and target domains, while the 26-65 classes were chosen as the unknown classes in the target domain. Moreover, we also constructed OSDA tasks between each two domains in both directions and formed 4$\times$3=12 tasks. For a fair comparison, we also used the deep features pre-extracted from the ResNet-50 model.

\subsection{Protocols}
\label{Protocols}
\par This paper proposes a newly-unified DA framework (i.e., IFCDA) to deal with challenges in both CSDA and OSDA scenarios. Therefore, we first report the classification results in CSDA paradigm, then we present the results in OSDA. As for the parameters involved in IFCDA, we fixed $T=5$ and $p=20$. However, we set different values for different cross-domain datasets referring to parameters $k$, $\gamma$, $\beta$, $\lambda$ and $\delta$ as follows: 1) USPS vs MNIST: $k=100$, $\gamma=0.1$, $\beta=1$, $\lambda=0.01$, $\delta=1$; 2) COIL-1 vs COIL-2: $k=20$, $\gamma=0.01$, $\beta=1$, $\lambda=0.01$, $\delta=1$; 3) MSRC vs VOC-2007: $k=20$, $\gamma=0.05$, $\beta=0.5$, $\lambda=0.01$, $\delta=0$; 4) Office-10 vs Caltech-10: $k=20$, $\gamma=0.05$, $\beta=0.5$, $\lambda=0.01$, $\delta=0.1$; 5) Office-Home: $k=100$, $\gamma=0.5$, $\beta=1$, $\lambda=0.1$, $\delta=1$.    

\par With respect to OSDA scenario, we set $k=100$, $\lambda=0.1$, $\delta=1$, $\gamma=1$ on both Office-31 and Office-Home datasets, where $\beta$ is no longer needed as we set $\textbf{\textit{A}}_s=\textbf{\textit{A}}_t$ for simplicity. Besides, this paper only focuses on more challenging unsupervised DA and there are no target labels, thus we set the parameters by searching a wide range and use the optimal ones exactly as previous work did \cite{IEEE:DICD,IEEE:JGSA}.

\begin{table*}[!t]
	\vspace*{-15pt}
	\fontsize{6}{5}\selectfont
	\renewcommand{\arraystretch}{1.3}
	\caption{AVERAGE CLASSIFICATION ACCURACY(\%) OF USPS VS MNIST IN CSDA SCENARIO}
	\label{table_2}
	\centering
	\begin{tabular}{|c||c|c|c|c|c|c|c|c|c|c|c|c|}
		\hline
		Tasks$/$Methods & GFK & SA & SDA & TCA & TJM & JDA & BDA & VDA & SCA & JGSA & DICD & IFCDA\\
		\hline
		\hline
		U$\rightarrow$M & 46.45 & 48.80 & 35.70 & 51.20 & 52.25 & 59.65 & 59.35 & 62.95 & 48.00 & \underline{68.15} & 65.20 & \textbf{73.40} \\
		\hline
		M$\rightarrow$U & 61.22 & 67.78 & 65.00 & 56.33 & 63.28 & 67.28 & 69.78 & 74.72 & 65.11 & \underline{80.44} & 77.83 & \textbf{85.72}\\
		\hline
		average         & 53.84 & 58.29 & 50.35 & 53.77 & 57.77 & 63.47 & 64.57 & 68.84 & 56.56 & \underline{74.30} & 71.52 & \textbf{79.56}\\
		\hline
	\end{tabular}
	\vspace{-10pt}
\end{table*}

\begin{table*}[!t]
	\fontsize{6}{5}\selectfont
	\renewcommand{\arraystretch}{1.3}
	\caption{AVERAGE CLASSIFICATION ACCURACY(\%) OF COIL-1 VS COIL-2 IN CSDA SCENARIO}
	\label{table_3}
	\centering
	\begin{tabular}{|c||c|c|c|c|c|c|c|c|c|}
		\hline
		Tasks$/$Methods & GFK & TCA & TJM & JDA & BDA & VDA & JGSA & DICD & IFCDA\\
		\hline
		\hline
		C-1$\rightarrow$C-2 & 72.50 & 88.47 & 91.67 & 89.31 & 97.22 & \underline{99.31} & 91.25 & 95.69 & \textbf{100.0}\\
		\hline
		C-2$\rightarrow$C-1 & 74.17 & 85.83 & 91.53 & 88.47 & 96.81 & \underline{97.92} & 91.25 & 93.33 & \textbf{99.44}\\
		\hline
		average             & 73.34 & 87.15 & 91.60 & 88.89 & 97.02 & \underline{98.62} & 91.25 & 94.51 & \textbf{99.72}\\
		\hline
	\end{tabular}
	\vspace{-10pt}
\end{table*}

\begin{table*}[!t]
	\fontsize{6}{5}\selectfont
	\renewcommand{\arraystretch}{1.3}
	\caption{AVERAGE CLASSIFICATION ACCURACY(\%) OF MSRC VS VOC-2007 IN CSDA SCENARIO}
	\label{table_4}
	\centering
	\begin{tabular}{|c||c|c|c|c|c|c|c|c|}
		\hline
		Tasks$/$Methods & GFK & SA & TCA & TJM & JDA & SCA & JGSA & IFCDA\\
		\hline
		\hline
		Ms$\rightarrow$Vo & 30.63 & 30.90 & 31.70 & 32.48 & 30.72 & 32.75 & \underline{33.20} & \textbf{40.13}\\
		\hline
		Vo$\rightarrow$Ms & 44.47 & 46.88 & 45.78 & 46.34 & 43.50 & \underline{48.94} & 48.86 & \textbf{63.91}\\
		\hline
		average           & 37.55 & 38.89 & 38.74 & 39.41 & 37.11 & 40.85 & \underline{41.03} & \textbf{52.02}\\
		\hline
	\end{tabular}
	\vspace{-10pt}
\end{table*}

\begin{table*}[!t]
	\fontsize{6}{5}\selectfont
	\renewcommand{\arraystretch}{1.3}
	\caption{AVERAGE CLASSIFICATION ACCURACY(\%) OF OFFICE-10 VS CALTECH-10 IN CSDA SCENARIO}
	\label{table_5}
	\centering
	\begin{tabular}{|c||c|c|c|c|c|c|c|c|c|c|c|c|c|}
		\hline
		Tasks$/$Methods & GFK & SA & SDA & TCA & TJM & JDA & BDA & VDA & SCA & PUnDA & JGSA & DICD & IFCDA\\
		\hline
		\hline
		C$\rightarrow$A & 46.03 & 49.27 & 49.69 & 45.82 & 46.76 & 45.62 & 44.89 & 46.14 & 43.74 & 50.10 & \underline{51.46} & 47.29 & \textbf{61.38}\\
		\hline
		C$\rightarrow$W & 36.95 & 40.00 & 38.98 & 31.19 & 38.98 & 41.69 & 38.64 & 46.10 & 33.56 & 41.70 & 45.42 & \underline{46.44} & \textbf{61.36}\\
		\hline
		C$\rightarrow$D & 40.76 & 39.49 & 40.13 & 34.39 & 44.59 & 45.22 & 47.77 & \underline{51.59} & 39.49 & 45.80 & 45.86 & 49.68 & \textbf{57.96}\\
		\hline
		A$\rightarrow$C & 40.69 & 39.98 & 39.54 & \underline{42.39} & 39.45 & 39.36 & 40.78 & 42.21 & 38.29 & 39.50 & 41.50 & \underline{42.39} & \textbf{47.64}\\
		\hline
		A$\rightarrow$W & 36.95 & 33.22 & 30.85 & 36.27 & 42.03 & 37.97 & 39.32 & \underline{51.19} & 33.90 & 42.50 & 45.76 & 45.08 & \textbf{55.59}\\
		\hline
		A$\rightarrow$D & 40.13 & 33.76 & 33.76 & 33.76 & 45.22 & 39.49 & 43.31 & \underline{48.41} & 34.21 & 40.30 & 47.13 & 38.85 & \textbf{52.23}\\
		\hline
		W$\rightarrow$C & 24.76 & \underline{35.17} & 34.73 & 29.39 & 30.19 & 31.17 & 28.94 & 27.60 & 30.63 & 36.50 & 33.21 & 33.57 & \textbf{41.23}\\
		\hline
		W$\rightarrow$A & 27.56 & 39.25 & 39.25 & 28.91 & 29.96 & 32.78 & 32.99 & 26.10 & 30.48 & \underline{42.40} & 39.87 & 34.13 & \textbf{45.41}\\
		\hline
		W$\rightarrow$D & 85.35 & 75.16 & 75.80 & 89.17 & 89.17 & 89.17 & 91.72 & 89.18 & \underline{92.36} & 85.20 & 90.45 & 89.81 & \textbf{94.90}\\
		\hline
		D$\rightarrow$C & 29.30 & 34.55 & 35.89 & 30.72 & 31.43 & 31.52 & 32.50 & 31.26 & 32.32 & \underline{38.90} & 29.92 & 34.64 & \textbf{37.93}\\
		\hline
		D$\rightarrow$A & 28.71 & 39.87 & 38.73 & 31.00 & 32.78 & 33.09 & 33.09 & 37.68 & 33.72 & \underline{40.30} & 38.00 & 34.45 & \textbf{47.29}\\
		\hline
		D$\rightarrow$W & 80.34 & 76.95 & 76.95 & 86.10 & 85.42 & 89.49 & \underline{91.86} & 90.85 & 88.81 & 83.20 & \underline{91.86} & 91.19 & \textbf{93.56}\\
		\hline
		average         & 43.13 & 44.72 & 44.53 & 43.26 & 46.33 & 46.38 & 47.15 & 49.03 & 44.29 & 48.86 & \underline{50.04} & 48.96 & \textbf{58.04}\\
		\hline
	\end{tabular}
	\vspace{-10pt}
\end{table*}

\par It is noteworthy that this paper proposes an importance filtered mechanism to obtain filtered soft labels for positive knowledge transfer, where parameters $N$ and $\alpha_{set}$ have to be estimated. However, $N$ is insensitive to different datasets and better performances could be achieved when we set $N=2,3,4$. The parameter $\alpha_{set}$ is required only in OSDA scenario, it could be also verified that desirable results could be expected when we set $\alpha_{set}=0.97,0.98$.

\par Followed by the experimental protocols in \cite{IEEE:JDA,IEEE:JGSA,IEEE:DICD}, in CSDA paradigm, we perform the proposed approach IFCDA on the datasets of USPS vs MNIST, COIL-1 vs COIL-2, MSRC vs VOC-2007, Office-10 vs Caltech-10, Office-Home. Moreover, several state-of-the art methods are evaluated for comparison, i.e., Geodesic Flow Kernel (GFK) \cite{IEEE:GFK}, Subspace Alignment (SA) \cite{IEEE:SA}, Subspace Distribution Alignment (SDA) \cite{IEEE:SDA}, Transfer Component Analysis (TCA) \cite{IEEE:TCA}, Transfer Joint Matching (TJM) \cite{IEEE:TJM}, Joint Distribution Adaptation (JDA) \cite{IEEE:JDA}, Balanced Distribution Adaptation (BDA) \cite{IEEE:BDA}, Visual Domain Adaptation (VDA) \cite{IEEE:VDA}, Scatter Component Analysis (SCA) \cite{IEEE:SCA}, Probabilistic Unsupervised Domain Adaptation (PUnDA) \cite{IEEE:IBDA1}, Joint Geometrical and Statistical Alignment (JGSA) \cite{IEEE:JGSA}, Graph Adaptive Knowledge Transfer (GAKT) \cite{IEEE:GAKT}, Domain Invariant and Class Discriminative (DICD) \cite{IEEE:DICD}, Deep Adaptation Network (DAN) \cite{IEEE:DAN}, Domain-Adversarial training of Neural Networks (DANN) \cite{IEEE:DANN}, Joint Adaptation Networks (JAN) \cite{IEEE:JAN}, Conditional Adversarial Domain Adaptation (CDAN) \cite{IEEE:CDAN}.

\par In OSDA scenario, we followed the same settings as previous OSDA work \cite{IEEE:STA_OSDA,IEEE:OSDA,IEEE:OSDA_B}, and the datasets of Office-31 and Office-Home are evaluated. Since the OSDA research is rarely reported, we compared our approach IFCDA with not only the latest OSDA methods, i.e., Assign and Transform Iteratively (ATI-$\lambda$) \cite{IEEE:OSDA}, Open Set Back-Propagation (OSBP) \cite{IEEE:OSDA_B}, Towards Open Set Deep Networks (OpenMax) \cite{IEEE:OpenMax}, Separate to Adapt (STA) \cite{IEEE:STA_OSDA}, but also some non-OSDA methods, i.e., DANN \cite{IEEE:DANN}, DAN \cite{IEEE:DAN}, Residual Transfer Networks (RTN) \cite{IEEE:RTN}, Unsupervised Domain Adaptation by Backpropagation (UDABP) \cite{IEEE:UDABP}.     

\subsection{Results and Discussions}
\label{Results and Discussions}

\par The classification results of different DA approaches on datasets of USPS vs MNIST, COIL-1 vs COIL-2, MSRC vs VOC-2007 and Office-10 vs Caltech-10 in CSDA scenario are given in Table \ref{table_2} $\sim$ Table \ref{table_5}, and our approach IFCDA outperforms other DA methods on all of 18 evaluations. The results of average accuracy of IFCDA in terms of different cross-domain datasets are 79.56\%, 99.72\%, 52.02\%, 58.04\%, which have 5.26\%, 1.10\%, 10.99\%, 8.00\% improvements compared with the best baselines JGSA, VDA, JGSA, JGSA, respectively.    

\par GFK and SA aimed to align source and target subspaces geometrically using infinite or fixed number of projections, while TCA, TJM, JDA, BDA established a statistical metric called MMD, to calibrate different distributions statistically in a shared subspace. SCA, VDA further respected the discriminative structure of data, either in source or target domains. SDA matched source and target distributions as well as their subspaces, where one projection and data mean/variance are adopted. Since MMD metric has shown its superior ability in measuring distributional distance and only using one projection would fail to take other structure information of data into account, PUnDA utilized two projections and MMD metric to reduce marginal distribution difference between the two domains. JGSA further diminished conditional distribution divergence, and preserved discriminative information of source domain. Besides, DICD respected discriminative structure of both source and target domains based on JGSA. 

\par However, the category-relevant losses about target domain in their models are usually formulated by pseudo hard labels (i.e., class-wise MMD, class scatter matrix), so that they are sensitive to quality of predicted labels due to no available target labels. From Table. \ref{table_5}, it can be observed that JSGA behaviors worse than PUnDA on some DA tasks although JGSA further considered conditional distribution difference using class-wise MMD. Moreover, DICD also performs more awful than JGSA on most DA tasks although DICD further employed class scatter matrix to respect discriminative structure of target domain. We speculate that those category-relevant losses they further considered have to utilize target pseudo hard labels, and their performances heavily depend on the quality of predicted target labels and distributional gaps between the two domains. From Table. \ref{table_5}, GAKT introduced the soft labels, but they adopted deep features of the FC-7 layer in the VGG-F model and ignored the discriminative structure of data, thus the performances are worst among them. For a fair comparison, we also utilize the soft labels only, while the features are extracted from ResNet-50 and the class-scatter matrix (i.e., IFCDA$_s$) is further reformulated. It could be observed that IFCDA$_s$ achieves largely improvements compared with GAKT, but IFCDA$_s$ still performs worse than IFCDA. We reckon that soft labels introduce too many small noisy probabilities. Differently, this paper devises an importance filtered mechanism to obtain filtered soft labels, then jointly reformulates the class-wise MMD and class scatter matrix. Therefore, negative transfer incurred by hard and soft labels is mitigated greatly, and the proposed approach could achieve best results among them. 

\par The results on Office-Home are shown in Table. \ref{table_6}. To evaluate the effectiveness of our method compared with Deep DA algorithms, we also report results of 4 end-to-end Deep DA models, i.e., DAN, DANN, JAN and CDAN. It can be observed that our approach achieves 3.4\% improvement against the best baseline CDAN. Compared with shallow DA, Deep DA integrates feature extraction and knowledge transfer into a shared network and achieves promising results. However, some techniques, which has been proven effective in domain adaptation, are hard to be implemented with deep structure. For example, the class-wise MMD can be easily optimized by matrix operations but it is very tricky in deep networks. Besides, the results on Office-Home verify that our method could be applicable to large-scale dataset and is able to achieve favorable results. Furthermore, our method generally runs faster than Deep ones since we use off-the-shelf features. 


\par With regard to OSDA paradigm, different approaches are evaluated on Office-Home and Office-31 datasets, and their results are illustrated in Table. \ref{table_7} $\sim$ Table. \ref{table_9}. Following previous work \cite{IEEE:OSDA,IEEE:OSDA_B,IEEE:STA_OSDA}, we employ 3 evaluation metrics: OS: normalized accuracy for all classes including the unknown as one class; OS$^\ast$: normalized accuracy only on known classes; UNK: the accuracy of unknown samples. In compared experiments, we study the OS accuracy on Office-Home dataset, while OS, OS$^\ast$ and UNK classification results are evaluated on Office-31 dataset. From the compared results, the OSDA-based approaches outperform better than non-OSDA methods DANN, DAN, RNT and UDABP, since the unknown classes in the target domain might bring negative transfer. Among the OSDA methods, the OS accuracy of our method on Office-Home is 72.4\%, which has 2.9\% improvement compared with the best baseline STA. The OS, OS$^\ast$ and UNK results on Office-31 dataset with Fc-7 features and ResNet-50 features are 86.4\%, 86.3\%, 87.0\% and 95.5\%, 96.3\%, 87.3\%, respectively, which further verifies that our approach not only mitigates the distributional shift between the two domains greatly, but also detects the unknown classes in the target domain correctly.
          
\begin{table}[!t]
	\fontsize{6}{5}\selectfont
	\renewcommand{\arraystretch}{1.3}
	\caption{AVERAGE CLASSIFICATION ACCURACY(\%) OF OFFICE-HOME IN CSDA SCENARIO}
	\label{table_6}
	\centering
	\begin{tabular}{|c||c|c|c|c|c|c|c|}
		\hline
		Tasks$/$Methods & DAN & DANN & JAN & CDAN & GAKT & IFCDA$_s$ & IFCDA\\
		\hline
		\hline
		Ar$\rightarrow$Cl & 43.6 & 45.6 & 45.9 & \underline{50.7} & 34.5 & 52.9 & \textbf{58.8}\\
		\hline
		Ar$\rightarrow$Pr & 57.0 & 59.3 & 61.2 & \underline{70.6} & 43.6 & 72.3 & \textbf{77.3}\\
		\hline
		Ar$\rightarrow$Rw & 67.9 & 70.1 & 68.9 & \underline{76.0} & 55.3 & 76.1 & \textbf{79.3}\\
		\hline
		Cl$\rightarrow$Ar & 45.8 & 47.0 & 50.4 & \underline{57.6} & 36.1 & 56.9 & \textbf{60.9}\\
		\hline
		Cl$\rightarrow$Pr & 56.5 & 58.5 & 59.7 & \underline{70.0} & 52.7 & 69.5 & \textbf{76.1}\\
		\hline
		Cl$\rightarrow$Rw & 60.4 & 60.9 & 61.0 & \underline{70.0} & 53.2 & 68.1 & \textbf{73.2}\\
		\hline
		Pr$\rightarrow$Ar & 44.0 & 46.1 & 45.8 & \underline{57.4} & 31.6 & 59.7 & \textbf{61.4}\\
		\hline
		Pr$\rightarrow$Cl & 43.6 & 43.7 & 43.4 & \underline{50.9} & 40.6 & 50.9 & \textbf{53.3}\\
		\hline
		Pr$\rightarrow$Rw & 67.7 & 68.5 & 70.3 & \underline{77.3} & 61.4 & 79.1 & \textbf{79.3}\\
		\hline
		Rw$\rightarrow$Ar & 63.1 & 63.2 & 63.9 & \textbf{70.9} & 45.6 & 69.2 & 69.6 \\
		\hline
		Rw$\rightarrow$Cl & 51.5 & 51.8 & 52.4 & \underline{56.7} & 44.6 & 56.8 & \textbf{58.5}\\
		\hline
		Rw$\rightarrow$Pr & 74.3 & 76.8 & 76.8 & \underline{81.6} & 64.9 & 81.8 & \textbf{83.0}\\
		\hline
		average           & 56.3 & 57.6 & 58.3 & \underline{65.8} & 47.0 & 66.1 & \textbf{69.2}\\
		\hline
	\end{tabular}
\end{table}

\begin{table}[!t]
	\vspace{-10pt}
	\fontsize{6}{5}\selectfont
	\renewcommand{\arraystretch}{1.3}
	\caption{AVERAGE CLASSIFICATION ACCURACY(\%) OF OFFICE-HOME IN OSDA SCENARIO (OS)}
	\label{table_7}
	\centering
	\begin{tabular}{|c||c|c|c|c|c|c|}
		\hline
		Tasks$/$Methods & DANN & ATI-$\lambda$ & OSBP & OpenMax & STA & IFCDA\\
		\hline
		\hline
		Ar$\rightarrow$Cl & 54.6 & 55.2 & 56.7 & 56.5 & \underline{58.1} & \textbf{61.9}\\
		\hline
		Ar$\rightarrow$Pr & 69.5 & 69.1 & 67.5 & 69.1 & \underline{71.6} & \textbf{74.3}\\
		\hline
		Ar$\rightarrow$Rw & 80.2 & 79.2 & 80.6 & 80.3 & \underline{85.0} & \textbf{87.6}\\
		\hline
		Cl$\rightarrow$Ar & 61.9 & 61.7 & 62.5 & 64.1 & \underline{63.4} & \textbf{66.5}\\
		\hline
		Cl$\rightarrow$Pr & 63.5 & 63.5 & 65.5 & 64.8 & \underline{69.3} & \textbf{69.3}\\
		\hline
		Cl$\rightarrow$Rw & 71.7 & 72.9 & 74.7 & 73.0 & \underline{75.8} & \textbf{79.0}\\
		\hline
		Pr$\rightarrow$Ar & 63.3 & 64.5 & 64.8 & 64.0 & \underline{65.2} & \textbf{69.7}\\
		\hline
		Pr$\rightarrow$Cl & 49.7 & 52.6 & 51.5 & 52.9 & \underline{53.1} & \textbf{54.3}\\
		\hline
		Pr$\rightarrow$Rw & 74.2 & 75.8 & 71.5 & 76.9 & \underline{80.8} & \textbf{83.9}\\
		\hline
		Rw$\rightarrow$Ar & 71.3 & 70.7 & 69.3 & 71.2 & \underline{74.9} & \textbf{77.8}\\
		\hline
		Rw$\rightarrow$Cl & 51.9 & 53.5 & 49.2 & 53.7 & \underline{54.4} & \textbf{58.6}\\
		\hline
		Rw$\rightarrow$Pr & 72.9 & 74.1 & 74.0 & 74.5 & \underline{81.9} & \textbf{85.9}\\
		\hline
		average           & 65.4 & 66.1 & 65.7 & 66.7 & \underline{69.5} & \textbf{72.4}\\
		\hline
	\end{tabular}
\end{table}

\begin{table}[!t]
	\vspace{-10pt}
	\fontsize{6}{5}\selectfont
	\renewcommand{\arraystretch}{1.3}
	\caption{AVERAGE CLASSIFICATION ACCURACY(\%) OF OFFICE-31, FC-7 FEATURES IN OSDA SCENARIO (OS, OS$^{\ast}$, UNK)}	
	\label{table_8}
	\centering
	\begin{tabular}{|c|c||c|c|c|c|c|c|}
		\hline
		Tasks & Metrics$/$Methods & DAN & RTN & UDABP & ATI-$\lambda$ & IFCDA\\
		\hline
		\hline
		\multirow{3}{*}{A$\rightarrow$D} & OS & 77.6 & 76.6 & 78.3 & \underline{79.8} & \textbf{80.4}\\
		\cline{2-7}
		& OS$^{\ast}$ & 76.5 & 74.7 & 77.3 & \underline{79.2} & \textbf{79.9}\\
		\cline{2-7}
		& UNK & - & - & - & \underline{85.8} & \textbf{86.0}\\
		\hline
		\multirow{3}{*}{A$\rightarrow$W} & OS & 72.5 & 73.0 & 75.9 & \underline{77.6} & \textbf{79.5}\\
		\cline{2-7}
		& OS$^{\ast}$ & 70.2 & 70.8 & 73.8 & \underline{76.5} & \textbf{78.6}\\
		\cline{2-7}
		& UNK & - & - & - & \underline{88.6} & \textbf{88.8}\\
		\hline
		\multirow{3}{*}{D$\rightarrow$A} & OS & 57.0 & 57.2 & 57.6 & \underline{71.3} & \textbf{76.6}\\
		\cline{2-7}
		& OS$^{\ast}$ & 53.5 & 53.8 & 54.1 & \underline{70.0} & \textbf{75.6}\\
		\cline{2-7}
		& UNK & - & - & - & \underline{84.3} & \textbf{87.0}\\
		\hline
		\multirow{3}{*}{D$\rightarrow$W} & OS & 88.4 & 89.0 & 89.8 & \underline{93.5} & \textbf{96.8}\\
		\cline{2-7}
		& OS$^{\ast}$ & 87.5 & 88.1 & 88.9 & \underline{93.2} & \textbf{97.3}\\
		\cline{2-7}
		& UNK & - & - & - & \textbf{96.5} & 91.6\\
		\hline
		\multirow{3}{*}{W$\rightarrow$A} & OS & 60.8 & 62.4 & 64.0 & \underline{76.7} & \textbf{86.3}\\
		\cline{2-7}
		& OS$^{\ast}$ & 58.5 & 60.2 & 61.8 & \underline{76.5} & \textbf{86.9}\\
		\cline{2-7}
		& UNK & - & - & - & \underline{78.7} & \textbf{81.0}\\
		\hline
		\multirow{3}{*}{W$\rightarrow$D} & OS & 98.3 & \textbf{98.8} & 98.7 & 98.3 & 98.5\\
		\cline{2-7}
		& OS$^{\ast}$ & 97.5 & 98.3 & 98.0 & \underline{99.2} & \textbf{99.6}\\
		\cline{2-7}
		& UNK & - & - & - & \textbf{89.3} & 87.6\\
		\hline
		\multirow{3}{*}{average} & OS & 75.8 & 76.2 & 77.4 & \underline{82.9} & \textbf{86.4}\\
		\cline{2-7}
		& OS$^{\ast}$ & 74.0 & 74.3 & 75.7 & \underline{82.4} & \textbf{86.3}\\
		\cline{2-7}
		& UNK & - & - & - & \textbf{87.2} & 87.0\\
		\hline
	\end{tabular}
	\vspace{-15pt}
\end{table}

\begin{table}[!t]
	\vspace{-10pt}
	\fontsize{6}{5}\selectfont
	\renewcommand{\arraystretch}{1.3}
	\caption{AVERAGE CLASSIFICATION ACCURACY(\%) OF OFFICE-31, RESNET-50 FEATURES IN OSDA SCENARIO (OS, OS$^{\ast}$, UNK)}	
	\label{table_9}
	\centering
	\begin{tabular}{|c|c||c|c|c|c|c|c|}
		\hline
		Tasks & Metrics$/$Methods & RTN & ATI-$\lambda$ & OSBP & STA & IFCDA\\
		\hline
		\hline
		\multirow{3}{*}{A$\rightarrow$D} & OS & 89.5 & 84.3 & 88.6 & \underline{93.7} & \textbf{94.9}\\
		\cline{2-7}
		& OS$^{\ast}$ & 90.1 & 86.6 & 89.2 & \underline{96.1} & \textbf{96.5}\\
		\cline{2-7}
		& UNK & - & 61.3 & \textbf{82.6} & 69.7 & 78.7\\
		\hline
		\multirow{3}{*}{A$\rightarrow$W} & OS & 85.6 & 87.4 & 86.5 & \underline{89.5} & \textbf{92.1}\\
		\cline{2-7}
		& OS$^{\ast}$ & 88.1 & 88.9 & 87.6 & \underline{92.1} & \textbf{92.7}\\
		\cline{2-7}
		& UNK & - & 72.4 & \underline{75.5} & 63.5 & \textbf{85.7}\\
		\hline
		\multirow{3}{*}{D$\rightarrow$A} & OS & 72.3 & 78.0 & 88.9 & \underline{89.1} & \textbf{96.0}\\
		\cline{2-7}
		& OS$^{\ast}$ & 72.8 & 79.6 & 90.6 & \underline{93.5} & \textbf{96.6}\\
		\cline{2-7}
		& UNK & - & 62.0 & \underline{71.9} & 45.1 & \textbf{90.4}\\
		\hline
		\multirow{3}{*}{D$\rightarrow$W} & OS & 94.8 & 93.6 & 97.0 & \underline{97.5} & \textbf{98.3}\\
		\cline{2-7}
		& OS$^{\ast}$ & 96.2 & 95.3 & \underline{96.5} & \underline{96.5} & \textbf{99.3}\\
		\cline{2-7}
		& UNK & - & 76.6 & \textbf{100.0} & \textbf{100.0} & 88.1\\
		\hline
		\multirow{3}{*}{W$\rightarrow$A} & OS & 73.5 & 80.4 & 85.8 & \underline{87.9} & \textbf{92.7}\\
		\cline{2-7}
		& OS$^{\ast}$ & 73.9 & 81.4 & 84.9 & \underline{87.4} & \textbf{92.7}\\
		\cline{2-7}
		& UNK & - & 70.4 & \textbf{94.8} & 92.9 & 93.5\\
		\hline
		\multirow{3}{*}{W$\rightarrow$D} & OS & 97.1 & 96.5 & 97.9 & \textbf{99.5} & 98.9\\
		\cline{2-7}
		& OS$^{\ast}$ & 98.7 & 98.7 & 98.7 & \underline{99.6} & \textbf{100.0}\\
		\cline{2-7}
		& UNK & - & 74.5 & 89.9 & \textbf{98.5} & 87.6\\
		\hline
		\multirow{3}{*}{average} & OS & 85.5 & 86.7 & 90.8 & \underline{92.9} & \textbf{\textit{95.5}}\\
		\cline{2-7}
		& OS$^{\ast}$ & 86.6 & 88.4 & 91.3 & \underline{94.2} & \textbf{96.3}\\
		\cline{2-7}
		& UNK & - & 69.5 & \underline{85.8} & 78.3 & \textbf{87.3}\\
		\hline
	\end{tabular}
	\vspace{-5pt}
\end{table}

\begin{table*}[!t]
	\fontsize{6}{5}\selectfont
	\renewcommand{\arraystretch}{1.3}
	\caption{PARAMETER SENSITIVITY of $N$ IN CSDA SCENARIO}
	\label{table_10}
	\centering
	\begin{tabular}{|c||c|c|c|c|c|c|c|c|c|c|}
		\hline
		Datesets$/$$N$ & 1 & 2 & 3 & 4 & 5 & 6 & 7 & 8 & 9 & All\\
		\hline
		\hline
		USPS vs MNIST & 78.9 & 79.3 & \textbf{79.5} & 79.4 & 79.4 & 79.3 & 79.3 & 79.3 & \textbf{79.5} & 79.2\\
		\hline
		C-1 vs C-2 & \textbf{99.7} & \textbf{99.7} & 99.5 & 99.1 & 99.0 & 99.2 & 99.2 & 99.2 & 99.2 & 98.5\\
		\hline
		MSRC vs VOC-2007 & 50.8 & 51.6 & \textbf{51.9} & 51.7 & 51.7 & - & - & - & - & 50.1\\
		\hline
		Office-10 vs Caltech-10 & 55.8 & 57.2 & 57.3 & 57.4 & \textbf{57.6} & \textbf{57.6} & 57.5 & 57.5 & 57.3 & 55.8\\
		\hline
		Office-Home & 68.8 & 68.7 & 68.8 & 68.9 & \textbf{69.0} & 69.0 & 68.9 & 68.9 & 68.9 & 66.1\\
		\hline
	\end{tabular}
	\vspace{-10pt}
\end{table*}

\begin{table*}[!t]
	\fontsize{6}{5}\selectfont
	\renewcommand{\arraystretch}{1.3}
	\caption{PARAMETER SENSITIVITY of $\alpha_{set}$ IN OSDA SCENARIO, RESNET-50 FEATURES (OS, OS$^{\ast}$, UNK)}	
	\label{table_11}
	\centering
\begin{tabular}{|c|c||c|c|c|c|c|c|c|c|c|c|c|c|}
	\hline
	Tasks & Metrics$/$$\alpha_{set}$ & 1.00 & 0.99 & 0.98 & 0.97 & 0.96 & 0.95 & 0.94 & 0.93 & 0.92 & 0.91 & 0.90\\
	\hline
	\hline
	\multirow{3}{*}{Ar$\rightarrow$Cl}&OS&                             71.8&70.2&\textbf{61.9}&55.7&46.1&38.9&35.2&30.7&26.8&23.0&19.7\\
	\cline{2-13}
	&OS$^{\ast}$&                                                      74.6&71.0&\textbf{61.3}&54.4&44.3&36.7&32.8&28.1&24.0&20.0&16.6\\
	\cline{2-13}
	&UNK&                                                              0.00&49.1&\textbf{76.8}&87.1&91.9&94.1&95.3&96.2&96.9&98.0&98.6\\
	\hline
	\multirow{3}{*}{Ar$\rightarrow$Pr}&OS&                             85.2&84.6&78.9&\textbf{74.3}&68.7&60.8&49.8&44.4&39.4&35.4&31.4\\
	\cline{2-13}
	&OS$^{\ast}$&                                                      88.6&86.1&79.2&\textbf{73.9}&67.9&59.5&48.0&42.3&37.0&32.9&28.7\\
	\cline{2-13}
	&UNK&                                                              0.00&46.7&70.1&\textbf{83.8}&89.0&91.4&94.2&95.7&96.9&97.5&97.7\\
	\hline
	\multirow{3}{*}{Cl$\rightarrow$Pr}&OS&                             77.5&78.1&\textbf{69.3}&65.6&61.3&56.0&49.8&43.7&37.4&33.0&29.9\\
	\cline{2-13}
	&OS$^{\ast}$&                                                      80.6&79.6&\textbf{69.3}&64.9&60.3&54.6&48.1&41.7&35.1&30.6&27.2\\
	\cline{2-13}
	&UNK&                                                              0.00&42.3&\textbf{70.4}&82.0&87.6&90.3&91.9&93.1&94.2&95.3&96.0\\
	\hline
	\multirow{3}{*}{Pr$\rightarrow$Cl}&OS&                             62.2&62.4&\textbf{54.3}&49.0&44.6&38.6&34.4&29.9&26.4&23.2&20.9\\
	\cline{2-13}
	&OS$^{\ast}$&                                                      64.7&63.1&\textbf{53.6}&47.8&43.0&36.6&32.2&27.5&23.8&20.4&18.0\\
	\cline{2-13}
	&UNK&                                                              0.00&45.7&\textbf{72.0}&80.6&84.9&87.7&89.1&90.4&91.6&92.4&93.2\\
	\hline
	\multirow{3}{*}{Rw$\rightarrow$Cl}&OS&                             68.7&69.1&63.8&\textbf{58.6}&51.5&46.9&42.8&39.2&36.5&34.1&30.8\\
	\cline{2-13}
	&OS$^{\ast}$&                                                      71.5&70.4&63.7&\textbf{57.8}&50.2&45.2&40.9&37.1&34.3&31.7&28.3\\
	\cline{2-13}
	&UNK&                                                              0.00&35.8&67.8&\textbf{78.1}&84.3&88.4&91.1&92.2&92.9&93.8&94.6\\
	\hline
	\multirow{3}{*}{Rw$\rightarrow$Pr}&OS&                             84.9&86.8&86.6&\textbf{85.9}&83.8&79.9&77.1&73.9&71.7&68.9&65.8\\
	\cline{2-13}
	&OS$^{\ast}$&                                                      88.3&88.7&87.7&\textbf{86.1}&83.7&79.5&76.5&73.1&70.7&67.8&64.6\\
	\cline{2-13}
	&UNK&                                                              0.00&38.6&59.8&\textbf{80.9}&85.4&89.1&91.6&93.7&94.5&95.4&96.0\\
	\hline
\end{tabular}
\vspace{-5pt}
\end{table*}

\begin{figure}[!t]
	\centering
	\includegraphics[width=0.8\linewidth,height=0.18\textheight]{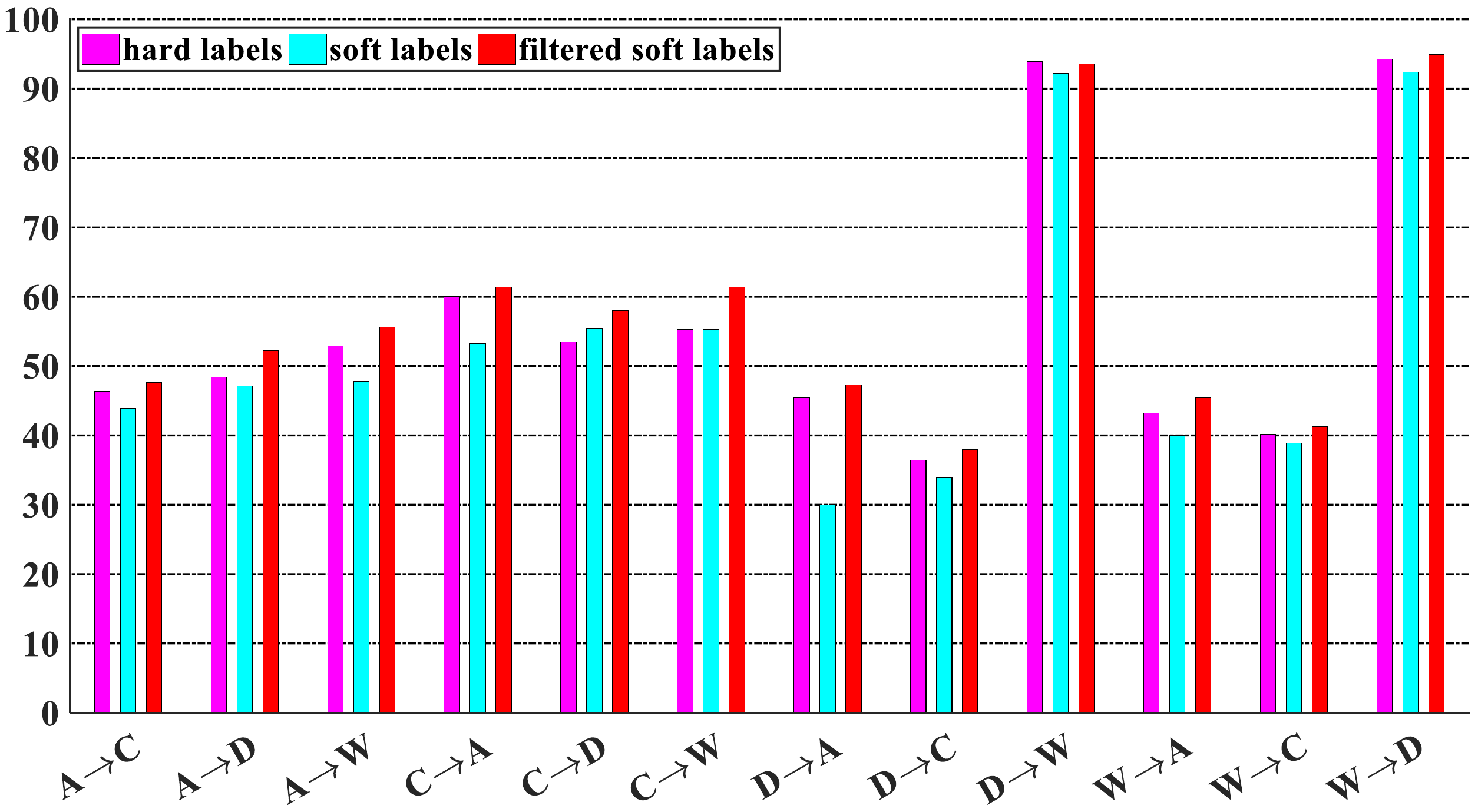}
	\caption{3 different labels are evaluated on Office-10 vs Caltech-10 in CSDA scenario.}
	\vspace{-15pt}
	\label{fig3}
\end{figure}

\begin{figure*}[!t]
	\centering
	\includegraphics[width=1.0\linewidth,height=0.5\textheight]{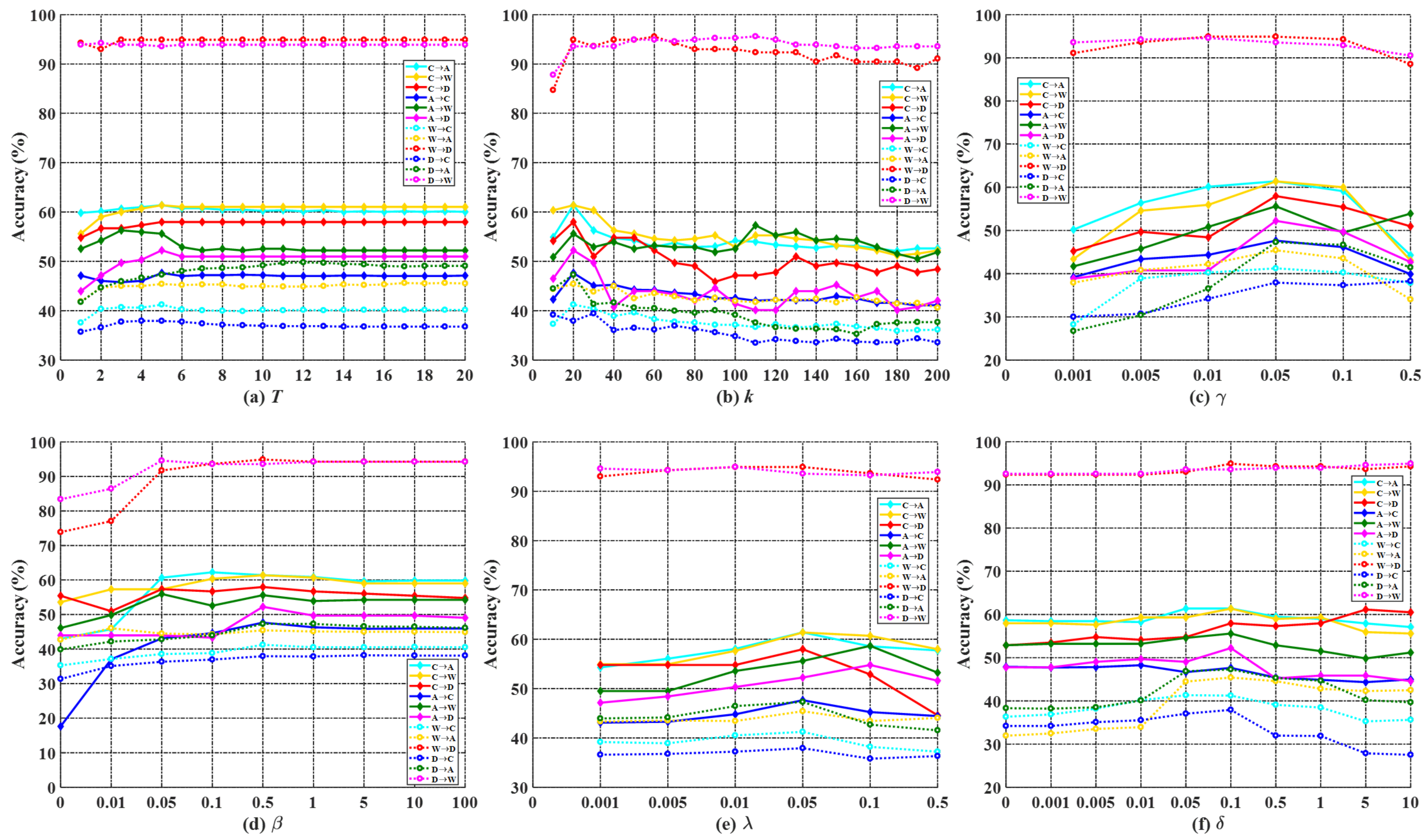}
	\caption{Parameter sensitivity of $T$, $k$, $\gamma$, $\beta$, $\lambda$ and $\delta$ values in CSDA.}
	\vspace{-15pt}
	\label{fig4}
\end{figure*}

\subsection{Model Analysis}
\label{Model Analysis}

\par In this section, we analysis effectiveness of the proposed approach as follows: 1) Superiority of filtered soft labels. 2) Parameter sensitivity of $N$ and $\alpha_{set}$ in the filtering mechanism. 3) Parameter sensitivity of $T$, $k$, $\gamma$, $\beta$, $\lambda$ and $\delta$.


\subsubsection{Superiority of Filtered Soft Labels}

It can be observed from Fig. \ref{fig3} and that the proposed model using 3 different labels, are evaluated on Office-10 vs Caltech-10 in CSDA scenario, we found that the proposed filtered soft labels have an significant edge over hard and soft ones. Notably, soft labels goes down largely on some DA tasks (e.g., C$\rightarrow$A, D$\rightarrow$A) compared with hard ones, and we speculated that too many negligible probabilities are involved in soft labels, so that the model is highly unstable and serious negative transfer is introduced. Moreover, the hard labels are overconfident due to pseudo target labels, while the proposed filtered soft labels not only fall away those confusing probabilities, but also retain the prominent ones. Therefore, the proposed importance filtered mechanism could achieve better performances than both hard and soft labels.      

\subsubsection{Parameter Sensitivity of $N$ and $\alpha_{set}$}

Although 2 newly additional parameters of $N$ and $\alpha_{set}$ are introduced into the proposed model, we observed that essential performances are achieved only on several parameter values from Table. \ref{table_10}. Relating to $N$, it is deduced that a large number of data instances just hesitate in several classes or definitely pertain to one class, thus its discrete value range is $[2,4]$. The $\alpha_{set}$ reflects capability of novel classes detection as introduced before, the larger $\alpha_l$ is, the worse its detection capability is, and vice versa. From Table. \ref{table_10}, the UNK is 0 when $\alpha_l$ is set to the largest value 1, and UNK is close to 1 when it is set to the smallest value 0.9. To obtain desirable results about those 3 metrics, it can be just set to 0.97 and 0.98.   

\subsubsection{Parameter Sensitivity of $T$, $k$, $\gamma$, $\beta$, $\lambda$ and $\delta$} 

We tested the proposed approach on 12 DA tasks established from Office-10 vs Caltech-10, with varying values of one parameter after fixing the others (i.e., $T=5$, $k=20$, $\gamma=0.05$, $\beta=0.5$, $\lambda=0.01$, $\delta=0.1$). Notably, similar trends on all other cross-domain datasets are not shown due to space limitations. As illustrated in Fig. \ref{fig4} (a) $\sim$ Fig. \ref{fig4} (f), we plot classification results w.r.t., their different values, and choose $T\in[1,20]$, $k\in[10,200]$, $\gamma\in[0.001,0.5]$, $\beta\in[0,100]$, $\lambda\in[0,0.1]$, $\delta\in[0,10]$. Specifically, Fig. \ref{fig4} (a) shows that the accuracy results decrease steadily with more iterations and converge within only 5 iterations. $k$ is selected such that the low-dimension embedding is accurate for original data reconstruction, and $\gamma$ is to control the scale of those two projections. Theoretically, $k$ and $\gamma$ are positively related to each other, since larger dimension and larger data scale has to be regulated. Therefore, their classification accuracy curves have similar feature of first rise up and then move down from Fig. \ref{fig4} (b), Fig. \ref{fig4} (c).

\par Concerning $\beta$, it dominates the deviation of source and target subspaces (i.e., $\textbf{\textit{A}}_s$, $\textbf{\textit{A}}_t$). As depicted in Fig. \ref{fig4} (d), when $\beta\rightarrow0$, the source subspace would be deviated infinitely from target one so that the feature transferability performs awfully. Conversely, with gradually increased value of $\beta$, the accuracy results would be slightly diminished, since the two subspaces would be aligned strictly and domain-specific structures of data damaged greatly. Additionally, $\lambda$ and $\delta$ embody the significance of specific domain-invariant discrimiantive structure of data and cross domain-invariant features respectively, and they are negatively related to each other. As illustrated in Fig. \ref{fig4} (e) and Fig. \ref{fig4} (f), their classification accuracy curves come down after first increase, since the other characteristic would be damaged when one property fully takes up the dominant position. Therefore, we better consider them simultaneously during knowledge transfer process.

\section{Conclusion}
\label{Conclusion}

\par In this paper, we argue that both the hard and soft labels adopted in DA are easily to cause negative transfer since the hard labels are overconfident and soft ones are ambiguous. Therefore, an importance filtered mechanism is proposed to fall away those negligible but confusing probabilities in soft labels, while the illustrious ones are reserved. To obtain soft labels, an approach of general graph-based label propagation is elaborated, which could realize label prediction in both CSDA and OSDA scenarios. Most importantly, some commonly used losses in DA, either category-relevant or category-irrelevant ones, are reformulated as more general forms using proposed filter soft labels, so that they could mitigate negative transfer incurred by hard and soft labels, and are versatile in that it could be used uniformly for both CSDA and OSDA scenarios.   


%

%

\section*{Acknowledgment}

This work was supported in part by the National Natural Science Foundation of China (NSFC) under Grants No.61772108, No.61932020, No.61976038, No.U1908210 and No.61976042.

\ifCLASSOPTIONcaptionsoff
  \newpage
\fi



%
%
%
\bibliographystyle{IEEEtran}
\bibliography{IEEETrans_our}

%

\vspace{-40pt}
\begin{IEEEbiography}[{\includegraphics[width=1in,height=1.25in,clip,keepaspectratio]{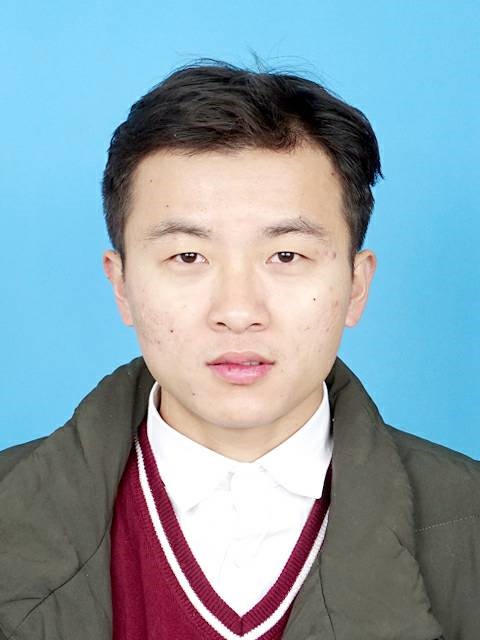}}]{Wei Wang}
is currently a Ph.D. candidate at the School of Software Technology, Dalian University of Technology, Dalian, China. He received the B.S. degree at the school of science from the Anhui Agricultural University, Hefei, China, in 2015. He received the M.S. degree at the school of computer science and technology from the Anhui University, Hefei, China, in 2018. His major research interests include  transfer learning, zero-shot learning, deep learning, etc.
\end{IEEEbiography}
\vspace{-40pt}
\begin{IEEEbiography}[{\includegraphics[width=1in,height=1.25in,clip,keepaspectratio]{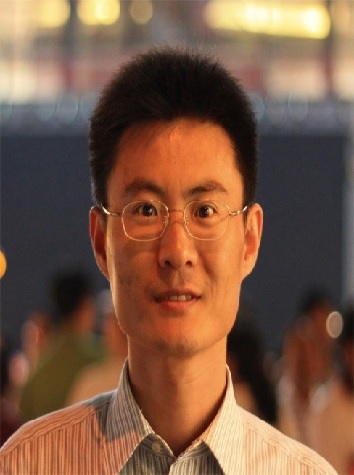}}]{Haojie Li}
	is currently a Professor in the DUT-RU International School of Information Science \& Engineering, Dalian University of Technology. He received the B.E. and the Ph.D. degrees from Nankai University, Tianjin and the Institute of Computing Technology, Chinese Academy of Sciences, Beijing, in 1996 and 2007 respectively. From 2007 to 2009, he was a Research Fellow in the School of Computing, National University of Singapore. He is a member of IEEE and ACM. His research interests include social media computing and multimedia information retrieval. He has co-authored over 70 journal and conference papers in these areas, including IEEE TCSVT, TMM, TIP, ACM Multimedia, ACM ICMR, etc.
\end{IEEEbiography}
\vspace{-40pt}
\begin{IEEEbiography}[{\includegraphics[width=1in,height=1.25in,clip,keepaspectratio]{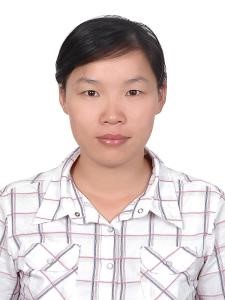}}]{Zhihui Wang}
received the B.S. degree in software engineering in 2004 from the North Eastern University, Shenyang, China. She received her M.S. degree in software engineering in 2007 and the Ph.D degree in software and theory of computer in 2010, both from the Dalian University of Technology, Dalian, China. Since November 2011, she has been a visiting scholar of University of Washington. Her current research interests include image processing and image compression.
\end{IEEEbiography}
\vspace{-40pt}
\begin{IEEEbiography}[{\includegraphics[width=1in,height=1.25in,clip,keepaspectratio]{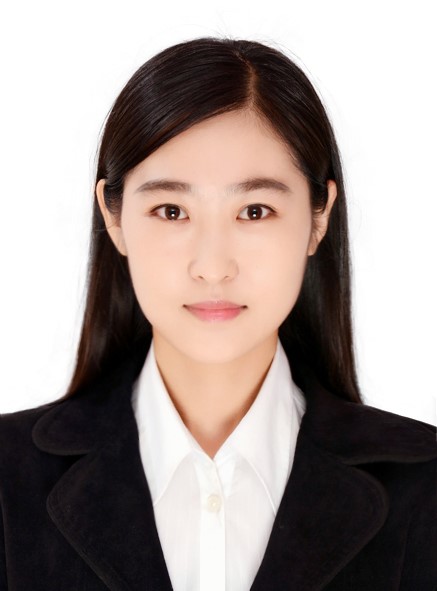}}]{Jing Sun}
	is currently a Ph.D. candidate at the School of Software Technology, Dalian University of Technology, Dalian, China. She received the M.S. degree in communication engineering from Liaoning University of Technology, Jinzhou, China, in 2017. Her major research interests include pattern recognition and transfer learning.
\end{IEEEbiography}
\vspace{-40pt}
\begin{IEEEbiography}[{\includegraphics[width=1in,height=1.25in,clip,keepaspectratio]{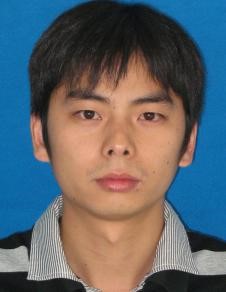}}]{Zhengming Ding}
	received the B.Eng. degree in information security and the M.Eng. degree in computer software and theory from University of Electronic Science and Technology of China (UESTC), China, in 2010 and 2013, respectively. He received the Ph.D. degree from the Department of Electrical and Computer Engineering, Northeastern University, USA in 2018. He is a faculty member affiliated with Department of Computer, Information and Technology, Indiana University-Purdue University Indianapolis since 2018. His research interests include transfer learning, multi-view learning and deep learning. He received the National Institute of Justice Fellowship during 2016-2018. He was the recipients of the best paper award (SPIE 2016) and best paper candidate (ACM MM 2017). He is currently an Associate Editor of the Journal of Electronic Imaging (JEI). He is a member of IEEE.
\end{IEEEbiography}
\vspace{-40pt}
\begin{IEEEbiography}[{\includegraphics[width=1in,height=1.25in,clip,keepaspectratio]{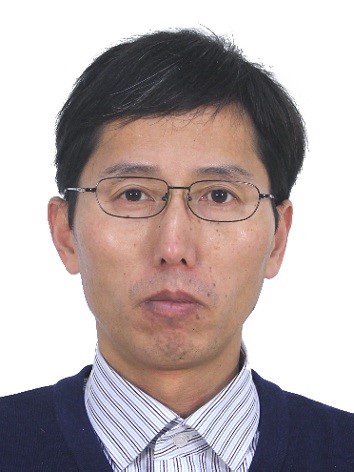}}]{Fuming Sun}
received the Ph.D. degree from the University of Science and Technology of China (USTC), Hefei, China, in 2007. From September 2012 to July 2013, he was a Visiting Scholar with the Department of Automation, Tsinghua University. He was a Professor with the School of Electronics and Information Engineering, Liaoning University of Technology, from 2004 to 2018. He is currently a Professor with the School of Information and Communication Engineering, Dalian Minzu University, Dalian, China. His current research interests include content-based image retrieval, image content analysis, and pattern recognition.
\end{IEEEbiography}






\end{document}